%% file: main.tex
\documentclass[letterpaper, 10 pt, conference]{ieeeconf}
\usepackage{amsmath}
\usepackage{amssymb} 
\usepackage{ntheorem}
\usepackage{cite}
\usepackage{bm} 
\usepackage[font=small]{caption}
\usepackage{graphicx}
\usepackage{subcaption}
\usepackage{comment} 
\usepackage{tabularx}
\newtheorem{Remark}{Remark}
\newtheorem{Problem}{Problem}
\newtheorem{Assumption}{Assumption}

\IEEEoverridecommandlockouts
\title{\LARGE \bf Impact-Robust Posture Optimization for Aerial Manipulation}

\author{Amr Afifi$^{*,1}$, Ahmad Gazar$^{*,2}$, Javier Alonso-Mora$^{2}$, Paolo Robuffo Giordano$^{3}$, Antonio Franchi$^{1,4}$%
\thanks{$^{*}$These authors contributed equally to this work.}
\thanks{This work was partially funded by the ANR-20-CE33-0003 ``CAMP'' project, the EU AUTOASSESS project, EU Grant agreement ID: 101120732. and by the Dutch Research Council NWO-NWA, within the “Acting under Uncertainty” (ACT) project (Grant No.NWA.1292.19.298))}
\thanks{
        $^{1}$Robotics and Mechatronics Department, Electrical Engineering,  Mathematics, and Computer Science (EEMCS) Faculty, University of Twente, 7500 AE Enschede, The Netherlands.  {\tt\footnotesize amrafifi142@gmail.com, schol@r-franchi.eu}.
}%
\thanks{
        $^{2}$Cognitive Robotics (CoR) Department, Delft University of Technology, 2628 CD Delft, The Netherlands. \tt\footnotesize{ahmad.gazar@pm.me; j.alonsomora@tudelft.nl}.
}%
\thanks{
        $^{3}$ CNRS, Univ Rennes, Inria, IRISA, Campus de Beaulieu, 35042 Rennes Cedex, France. E-mail: {\tt\scriptsize prg@irisa.fr}
}%
\thanks{
        $^{4}$Department of Computer, Control and Management Engineering, Sapienza University of Rome, 00185 Rome, Italy. {\tt\footnotesize schol@r-franchi.eu}
}%
}
\input{symbol_def}

\begin{document}

\maketitle

\begin{abstract}
We present a novel method for optimizing the posture of kinematically redundant torque-controlled robots to improve robustness during impacts. A rigid impact model is used as the basis for a configuration-dependent metric that quantifies the variation between pre- and post-impact velocities. By finding configurations (postures) that minimize the aforementioned metric, \emph{spikes} in the robot's state and input commands can be significantly reduced during impacts, improving safety and robustness. The problem of identifying impact-robust postures is posed as a min–max optimization of the aforementioned metric. To overcome the real-time intractability of the problem, we reformulate it as a gradient-based motion task that iteratively guides the robot towards configurations that minimize the proposed metric. This task is embedded within a task-space inverse dynamics (TSID) whole-body controller, enabling seamless integration with other control objectives. The method is applied to a kinematically redundant aerial manipulator performing repeated point contact tasks. We test our method inside a realistic physics simulator and compare it with the nominal TSID. Our method leads to a reduction (up to 51\% w.r.t. standard TSID) of post-impact spikes in the robot's configuration and successfully avoids actuator saturation. Moreover, we demonstrate the importance of kinematic redundancy for impact-robustness using additional numerical simulations on a quadruped and a humanoid robot, resulting in up to 45\%, reduction of post-impact spikes in the robot's state w.r.t. nominal TSID.  
\end{abstract}

\input{intro}

\input{preliminary_impact}

\input{theory_impact}
\input{application}
\input{simulations}
\input{legged_locomotion}

\input{discussion}
\input{conc}

\bibliographystyle{IEEEtran}
\bibliography{biblio}

\end{document}

%% file: symbol_def.tex
\usepackage{subcaption}
\usepackage{color}

\newcommand{\bnu}{\bm{\nu}}     
\newcommand{\bnudt}{\dot{\bm{\nu}}}     
\newcommand{\q}{\bm{q}}

\newcommand{\jacobc}{\bm{J}_{c}(\bm{q})}

\newcommand{\ip}{\bm{u}}

%% file: intro.tex
\section{Introduction}
The last two decades have witnessed significant progress in the field of Aerial Manipulation~\cite{9462539}. 
A key industrial application is Non-Destructive Testing (NDT), where aerial manipulators (AMs) are required to perform multiple point-contact inspections in a fast manner, reducing asset downtime and providing significant economic benefits~\cite{11072833}. However, for robotic systems, including AMs, fast making and breaking contact remains a challenging manipulation skill. In particular, impacts—contacts that occur at non-zero speed—induce a sudden change of momentum over a short time scale (milliseconds). If not handled properly, impact events can potentially lead to \emph{spikes} in robot actuator commands and in extreme cases to even actuator saturation~\cite{9636094}. Such phenomena can significantly compromise both the safety and robustness of the robot during impact events.

Operational space control is widely used for motion and force control of robotic systems~\cite{khatib1987}. Task Space Inverse Dynamics (TSID) extends this framework by solving a Quadratic Program (QP) with multiple prioritized tasks in the operational space while respecting the robot's full dynamics and constraints, making it well-suited for kinematically redundant robots~\cite{sentis2005, Righetti2011, Mistry2012, DelPrete2015, wensing2024}. However, in its basic formulation, TSID is impact-unaware, as it assumes smooth contact dynamics and does not explicitly model the instantaneous velocity changes caused by impacts. These changes are typically managed through high-frequency feedback loops, which may be insufficient for handling impacts. 

The question of how to deal with impacts effectively has been studied in the robotics community, with a focus on fixed-base manipulators and legged robots. In~\cite{7989472,van2024quadratic}, a control technique known as reference spreading is proposed. Reference spreading smooths out the reference target used in the feedback controller by “spreading” signal discontinuities over a short time window of the impact event. A limitation of this approach is that the tracking performance of the controller can degrade if the impact time is uncertain. Another approach in~\cite{9636094} defines a subspace of generalized velocities that remains unaffected by any contact impulse. Feedback control objectives (e.g., velocity tracking errors) are projected onto this impact‑invariant subspace, which results in the controller \emph{ignoring} parts of the state that are unpredictable due to impact, while still exerting full authority over the invariant dynamics. Although this approach is robust to any possible contact impulse, it can lead to overly conservative behavior. In~\cite{Kheddar2023}, an impact‑aware TSID formulation is proposed that explicitly models sudden state jumps (in joint velocity and torque) due to impacts. In this QP formulation, constraints are adjusted to anticipate these jumps within robot hardware limits, preventing controller infeasibility post-impact. However, one limitation of this approach lies in the challenge to accurately identify the post-impact velocity and torque bounds without being overly conservative, and it does not minimize the impact effect. The method closest to our proposed approach is presented in~\cite{walker1994},  where kinematic redundancy is exploited to optimize robot configurations for improved impact handling via differential inverse kinematics and a projected gradient approach. However, this is only suitable for velocity-controlled robots and does not allow for multiple task objectives and the inclusion of hard constraints.

Impact-handling approaches are still relatively unexplored in aerial manipulation contexts~\cite{indukumar2025study}. For AMs, impacts have been addressed by leveraging energetic concepts, for example, in~\cite{9812342}, where a passivity-based framework including energy-tanks is used to enable safe interaction with any passive environment. Impacts have also been dealt with via hardware designs, as in ~\cite{10557063,7387723}. 

In this work, we tackle impacts by exploiting kinematic redundancy for improving a robot's \textbf{impact-robustness}, that is, to reduce the risk of spikes in both the state and control commands and consequently avoid actuator saturation. Inspired by \cite{walker1994}, we adopt a rigid impact model to derive a configuration-dependent metric that quantifies the robot’s susceptibility to impacts. We propose a whole-body TSID controller that includes the search for an impact-robust posture within its cost objective. Different from~\cite{walker1994}, our QP-based approach is suitable for torque-controlled robots and allows for multiple task objectives and hard constraints applied to complicated, highly dimensional robotic systems. Our method is tailored for anticipated impact tasks, i.e., contact tasks where impacts are expected to occur. Consequently, the robot can assume a "brace for impact" posture before the actual impact. Our results show that our method effectively optimizes postures that reduce the risk of actuator saturation and lead to safer and more robust performance during impacts. Since our approach achieves impact-robustness via redundant degrees of freedom (DoF), our method is not conservative and does not compromise the task-space feedback control action as in~\cite{7989472,van2024quadratic,9636094}. Moreover, our approach does not require the identification of post-impact state bounds as in~\cite{Kheddar2023}, which can affect the controller's feasibility. 

Our contributions can be summarized as follows:
\begin{itemize}
    \item We propose a novel and generic method to improve robustness during impacts by optimizing the postures of kinematically redundant torque-controlled robots. The optimization is formulated as a motion task within a TSID whole-body QP-controller.
    \item Using a kinematically redundant AM, We test and validate our method in a realistic physics simulator on multiple contact NDT-like scenarios. Our method leads to a significant reduction of post-impact robot motion, w.r.t. nominal TSID, up to 51\% and avoids actuator saturation during impact events.
    \item We confirm the generality of the approach in a multi-contact, high kinematic redundancy setting of legged robots, by incorporating the impact-robustness task in a whole-body controller tailored for legged robots. The validity of the approach is confirmed via numerical simulations on a humanoid and quadruped robot. Again, our method results in a reduction of post-impact motion in the configuration (up to 45\% w.r.t. nominal TSID). 
\end{itemize}


%% file: preliminary_impact.tex
\section{Preliminaries}\label{sec:preliminaries}
In this section, we define our system dynamics modeling and present the feedback control challenges that robots face when interacting physically with the environment. 
\subsection{Floating Base Robot Dynamics}
The robots considered in this work are modeled as floating-base systems. The robot configuration is $\q = (\q^\mathcal{I}_b, \q_j) \in \mathcal{M} = \mathbb{SE}(3) \times \mathbb{R}^{n_j}$, where $\q^\mathcal{I}_b$ represents the floating-base pose, expressed in the inertial frame $\mathcal{I}$, while $\q_j$ denotes the joint positions of the robot, with $n_j$ Degrees of Freedom (DoF). The generalized robot velocities are denoted as $\bnu \in \mathbb{R}^{6+n_j}$. The equations of motion of such a system can be derived using the Euler-Lagrange equations of motion as~\cite{featherstone2007}: 
\begin{IEEEeqnarray}{LLL}
    \bm{M}(\q) \bnudt + \bm{h}(\q,\bnu) = \bm{G}(\q)\ip + \jacobc^{T}\bm{F}_c.
    \label{eq:realdyn}
\end{IEEEeqnarray}
The inertia matrix is denoted as $\bm{M}(\bm{q}) \in \mathbb{R}^{(6+n_j) \times (6+n_j)}$, and $\bm{h}(\boldsymbol{q}, \dot{\boldsymbol{q}}) \in \mathbb{R}^{6+n_j}$ is the vector capturing the Coriolis, centrifugal, and gravity effects. $\bm{J}_{c}$ represents the contact Jacobian associated with the generalized contact forces $\bm{F}_{c}$. 
\subsection{Aerial Manipulator Actuation Model}
The AM considered in this work is composed of a fully actuated hexarotor with fixedly tilted propellers and a $3$-DoF robotic arm attached to its base. The AM can be modeled as a floating base system as in Eq.~\eqref{eq:realdyn} where $n_j = 3$. 
The actuated control commands are denoted as $\small \bm{u} = \begin{bmatrix} \bm{u}_b^\top, \bm{\tau}^\top \end{bmatrix}^\top$, where $\bm{u}_b \in \mathbb{R}^6$ represent the propellers' angular velocities of the hexarotor, and $\bm{\tau} \in \mathbb{R}^3$ being the manipulators' joint torques. Each propeller angular velocity $\omega_i, \,\, \forall i= 1, ..., 6$ is related to its thrust and drag moment, in good approximation, quadratically. For more details, we refer the interested reader to ~\cite{hamandi2021design}. The AM control inputs are then mapped to the state tuple using the actuation matrix $\bm{G}(\q) \in \mathbb{R}^{n\times m}$.
\subsection{Whole Body Control as a QP}
Whole-body control of kinematically redundant robots can be formulated using TSID~\cite{sentis2005, Righetti2011, Mistry2012, DelPrete2015}. Given the current state of the robot ($\bm{q}$ and $\bm{\nu}$), the control objective is to find the set of controls $\bm{u}$, and joint accelerations $\dot{\bm{\nu}}$, that minimize the residual of a stack of tasks in the operational space subject to the dynamics and physical constraints of the robot. This problem can be formulated as a QP to achieve real-time instantaneous feedback control as follows:
\begin{Problem} Vanilla TSID
\label{eq:vanilla TSID problem}
\begin{subequations}    
\begin{IEEEeqnarray}{LLL}
    \underset{\bm{z}:=(\bm{u},\,\bm{\dot{\nu}})}{\text{minimize}} \sum^{N_{\text{tasks}}}_{i=1} w_i||\bm{e}_i(\bm{z})||^2   \label{eq:cost_tsid}
    \\
    \textrm{subject to} \nonumber
    \\
    \qquad \bm{M}(\bm{q}) \bnudt + \bm{h}(\bm{q}, \bm{\nu}) = \bm{G}(\bm{q}) \ip + \jacobc^{T}\bm{F}_c, \label{eq:dynamics}
    \\
    \qquad  \underline{\bm{{u}}} \leq \bm{u} \leq \overline{\bm{{u}}},\quad   \underline{\bm{{\dot{\nu}}}} \leq \bm{\dot{\nu}} \leq \overline{\bm{{\nu}}}. 
\label{eq:box constraints}  
\end{IEEEeqnarray}
\end{subequations}
\end{Problem}
The total optimizers are denoted as $\bm{z}$ (acceleration and controls). The cost objective~\eqref{eq:cost_tsid} is the summation of the weighted $l_2$ norm of the task acceleration residual $\bm{e}_i := \ddot{\bm{y}}_i - \ddot{\bm{y}}^\ast_i$, where $w_i,\, \forall i = 1,\dots, N_{\text{tasks}}$ is the $i$th task weight, and 
\begin{subequations}    
\begin{IEEEeqnarray}{LLL}
    \ddot{\bm{y}} &= \bm{J}_i (\bm{q}) \bnudt + \dot{\bm{J}}_i (\bm{q}) \bm{\nu}\label{eq:task_acceleration},
    \\
    \ddot{\bm{y}}^\ast_i &= \underbrace{\ddot{\bm{y}}_{\text{des}_i}}_{\bm{u}_{\text{ffwd}}} + \underbrace{\bm{K}_p(\bm{y}_{\text{des}_i} - \bm{y}) + \bm{K}_d(\dot{\bm{y}}_{\text{des}_i} - \dot{\bm{y}})}_{\bm{u}_\text{fdbk}}. \label{eq:desired_task_acceleration}
    \label{eq:vanilla_error}
\end{IEEEeqnarray}
\end{subequations}
The output task acceleration $\ddot{\bm{y}} \in \mathbb{R}^{n_t}$ in~\eqref{eq:task_acceleration} is achieved using the desired output-feedback controller $\ddot{\bm{y}}^\ast_i$ in~\eqref{eq:desired_task_acceleration} comprising a feedforward acceleration command $\ddot{\bm{y}}_{\text{des}_i}$, and a feedback control action with $\bm{K_p}\succ 0$, and $\bm{K_d} \succ 0$ being the positive definite position and velocity task feedback gains respectively. $\bm{J}_i(\bm{q}) \in \mathbb{R}^{n_t \times n}$ represents the task Jacobian that maps the joint velocities to the operational space velocity, with  $n_t$ being the task DoF. Finally,~\eqref{eq:box constraints} represent the lower and upper bounds on the controls and joint accelerations, respectively.
\begin{Remark}
\label{remark:weighted_tsid}
The task residuals in problem~\eqref{eq:vanilla TSID problem} in the cost objective are realized in a weighted fashion. An alternative approach can be enforcing a strict hierarchy between tasks, where lower-priority tasks are projected into the null space of the higher-priority tasks~\cite{sentis2005}. This hierarchical approach can be particularly advantageous for our AM application, as will be presented in the subsequent sections.
\end{Remark}
%
\subsection{Feedback Control Challenges during Impacts} 
\label{sec:control challenges}
Feedback control of robots in contact with the environment is a challenging task due to the fast, discontinuous nature of impacts that occur in orders of milliseconds~\cite{stewart1996implicit}. We summarize this challenge here for completeness, as highlighted in~\cite{9636094} as a motivation for our approach. Assuming a robot makes contact with a rigid body, the contact is established instantaneously (i.e., the robot's configuration $\bm{q}$ remains constant during the impact event, assuming no contact slippage). Simultaneously, there is an instantaneous change in the robot's velocity due to the impulse $\bm{\lambda}$. Mathematically, this rigid impact model can be written as follows: 
\begin{subequations}   
\label{eq:impact_model}
\begin{IEEEeqnarray}{LLL}
    \bm{q}^+ = \bm{q}^-
    \\
    \bnu^{+} - \bnu^{-}  = \bm{M}^{-1}(\q) \jacobc^{T} \bm{\lambda},
\end{IEEEeqnarray}
\end{subequations}
where, $\bm{q}^{+}$, $\bm{q}^{-}$,  and $\bnu^{+}$, $\bnu^{-}$ are the post- and pre-impact configurations and velocities respectively. Consider the following PD feedback controller similar to~\eqref{eq:desired_task_acceleration}:  
\begin{IEEEeqnarray}{LLL}    
    \bm{u}_{\text{fdbk}} = \bm{K}_p \bm{e}_p + \bm{K}_d \bm{e}_{\bm{\nu}},  
\end{IEEEeqnarray}
where $\bm{e}_p$, and  $\bm{e}_{\bm{\nu}}$ are the position and velocity errors simultenously. According to the rigid impact model assumption in~\eqref{eq:impact_model}, the magnitude of the feedback command during the impact is dominated by the change in velocity error since the configuration change is almost zero. This implies the following variational relationship at the impact event:
\begin{IEEEeqnarray}{LLL}    
    \Delta \bm{u}_{\text{fdbk}} \approx \bm{K}_d \Delta \bm{e}_{\bm{\nu}}. 
\end{IEEEeqnarray}

In simple words, large changes in the velocity feedback error exhibit large feedback residuals in problem~\eqref{eq:vanilla TSID problem}, which in turn lead to large control commands in milliseconds duration (spikes) that can potentially be infeasible for the robot actuators. This can degrade the controller's performance and hinder the robot's safety. The above analysis suggests that feedback alone may not suffice for controlling robots during impacts. We highlight that this analysis is generally applicable to feedback controllers during impact events, whether in joint space or operational space. In the following sections, we propose a method that demonstrates how to incorporate anticipated impacts into feedforward control commands to enhance the robustness of reactive controllers, particularly in kinematically redundant, optimization-based TSID controllers.

%% file: theory_impact.tex
\section{Method: Optimizing Impact-Robust Postures}\label{sec:method}
We aim to minimize the variation between the pre- and post-impact velocities, resulting in smoother control commands and improved robustness during impact events. 
\begin{Assumption}[Anticipated bounded impacts]
\label{assumption:impacts}
We assume anticipated impacts, i.e., impacts that are part of a contact task and for which the prospective contact Jacobian $\bm{J}_c(\bm{q})$ and its contact normal $\bm{n}$ are known a priori, with the impact magnitude $\lambda$ along $\bm{n}$ being a bounded uncertainty $|\lambda|\leq\overline{\lambda}$. 
\end{Assumption}
Given Assumption~\ref{assumption:impacts}, expanding the impulse vector at the velocity jump with its magnitude and direction becomes:
\begin{subequations}
\begin{IEEEeqnarray}{LLL}
    \Delta \bm{\nu} &= \bm{M}^{-1}(\bm{q}) \bm{J}_c^\top \bm{n}\,\lambda
    =: \bm{\Gamma}(\bm{q})\,\bm{n}\,\lambda,
    \label{eq:impact_velocity_residual} \\
    \bm{\Gamma}(\bm{q}) &\triangleq \bm{M}^{-1}(\bm{q}) \bm{J}_c^{\top}.
    \label{eq:impact_metric}
\end{IEEEeqnarray}
\end{subequations}
To minimize the variation between pre- and post-impact velocities, we formulate the following robust optimization problem:
\begin{IEEEeqnarray}{LLL}
\min_{\bm{q}}\; \max_{|\lambda|\leq\overline{\lambda}} 
\;\; \|\Delta \bm{\nu}\|^{2},
\label{eq:opt_minmax}
\end{IEEEeqnarray}
Substituting \eqref{eq:impact_velocity_residual}, we obtain
\begin{IEEEeqnarray}{LLL}    
\|\Delta \bm{\nu}\|^{2} = \lambda^2\, \bm{n}^\top \underbrace{\big(\bm{J}_c \bm{M}^{-2} \bm{J}_c^\top\big)}_{\bm{\Lambda}_c(\bm{q})}\bm{n},
\end{IEEEeqnarray}
where $\bm{\Lambda}_c=(\bm{J}_c \bm{M}^{-2} \bm{J}_c^\top)$ is known as the \emph{Dynamic Impact Ellipsoid}~\cite{walker1994}, which is a form of the effective inertia in the operational-space. The shape of this ellipsoid, computed using its SVD decomposition, determines how much the robot's inertia varies in the contact space. Since the worst-case $\lambda$ satisfies $|\lambda|=\overline{\lambda}$, the min–max problem reduces to
\begin{IEEEeqnarray}{LLL}
\label{eq:minimax ocp}
\min_{\bm{q}} \quad 
\overline{\lambda}^{2}\, \bm{n}^\top \bm{\Lambda}_c(\bm{q}) \bm{n}.
\label{eq:opt_min}
\end{IEEEeqnarray}
The above optimization problem chooses a posture $\bm{q}$ that maximizes the effective inertia along the impact direction. In other words, the robot is required to adjust its configuration to absorb impacts with larger body inertia in the impact direction.

Problem~\eqref{eq:opt_min} is a nonlinear optimization problem and can be intractable to solve online. Instead, we propose approximating it locally as a low-priority posture task within TSID. Define 
\begin{subequations}
\begin{IEEEeqnarray}{LLL}
H(\bm{q}) &:= \bm{n}^\top \bm{\Lambda}_c(\bm{q}) \bm{n}, 
\\
\bm{e}_{\text{IM-posture}} &=  \dot{\bm{\nu}} - \dot{\bm{\nu}}^\ast_{\text{IM-posture}}, 
\\
\dot{\bm{\nu}}^\ast_{\text{IM-posture}} &\triangleq 
 -\bm{K}\nabla_{\bm{q}} H(\bm{q})+ \bm{K}_p(\q_\text{des} - \q)-\bm{K}_d\bm{\nu},
\label{eq:desired_posture_acc}
\end{IEEEeqnarray}
\end{subequations}
where $\q_{\text{des}}$ is a desired robot configuration. $\bm{K}\succ0$ is a tunable gain. We call $H(\q) \in \mathbb{R}$ the \emph{Impact-Robustness Metric}, which quantifies the effect of impacts along a specific direction on the robot's effective inertia. The higher this number is, the higher the robot's effective inertia in that direction. Finally, $\nabla_{\bm{q}} H(\q)$ is defined as the gradient of the impact-robustness metric with respect to $\q$. 
 
The weighted $l_2$ norm for the impact-robustness task residual $\bm{e}_{\text{IM-posture}}$ can be incorporated within TSID as part of the objective function in Eq.~\ref{eq:cost_tsid}, in addition to an appropriate weighting choice that represents its priority. The proposed task residual will iteratively reconfigure the robot's posture to achieve the local minimization of ${H}(\q)$. Since we consider impacts which are part of a contact task, the robot can choose, e.g., based on a finite state machine, to activate this task residual (introduce a non-zero weight) before attempting the impact action. This highlights the flexibility of the method in being applied only when required by the task

%% file: application.tex
\section{Application to Aerial Manipulation}\label{sec:apply}
In this section, we apply the proposed method to an AM. We focus on inspection tasks where AMs are required to make fast repeated contacts with the environment.

\subsection{Aerial Manipulator Stack of Tasks}
As mentioned in remark~\ref{remark:weighted_tsid}, different from problem~\eqref{eq:vanilla TSID problem}, we adapt a strict hierarchy of tasks inside TSID that prioritizes the reduced-attitude (tilt) control of the fully-actuated multi-rotor as the highest priority task, the EE as a second priority, and the robot's posture as the third priority. Unlike earlier works on the control of fully actuated kinematically redundant AMs \cite{ryll2018truly} \cite{8928943}, we prioritize the tilt angle control, as it has a strong influence on the robot's stability. 
\subsubsection{Reduced Attitude Task}
\label{subsection:priority 1}
We assign a desired tilt of the hexarotor base as a first priority task. This is achieved using the control command $\dot{\bm{\omega}}_{xy}^{*} \in \mathbb{R}^2$, specifying the angular acceleration of the hexarotor around its tilt axis (i.e., the x-y directions). The desired angular acceleration is then related to the joint accelerations $\bnudt$ through the Jacobian task matrix $\bm{J}_{\text{tilt}} \in \mathbb{R}^{2 \times 6+n_j} $ as: 
\begin{subequations}    
\begin{IEEEeqnarray}{LLL}
\dot{\bm{\omega}}_{xy}^{*} &= \bm{J}_{\text{tilt}} \dot{\bm{\nu}},
\\
\bm{J}_{\text{tilt}} &\triangleq 
    \begin{bmatrix}
    0 & 0 & 0 & 1 & 0 & 0 & 0 & 0 & 0 
    \\
    0 & 0 & 0 & 0 & 1 & 0 & 0 & 0 & 0 
    \end{bmatrix}.
\end{IEEEeqnarray}
\end{subequations}
The reduced attitude error $\bm{e}_{\text{red}} \in \mathbb{R}^{2}$ is a measure of the error between the aerial robot's body frame's z-axis and the z-axis of the desired frame which is calculated based on the shortest rotation that aligns both z-axes \cite{brescianini2018tilt}.
The desired reduced attitude controller is then calculated as 
\begin{IEEEeqnarray}{LLL}
    \dot{\bm{\omega}}_{xy}^{*} = \bm{K}_{\text{red}} \bm{e}_{\text{red}} - \bm{K}_{\omega}\bm{\omega_{xy}}.
\end{IEEEeqnarray}%
\subsubsection{End-Effector Pose Task} 
\label{subsection:priority 2}
The second priority task controls the EE pose. We consider a desired EE impedance command ~\cite{4788393} $\bnudt_e^* \in \mathbb{R}^{6}$, with the following task kinematics expressed at the acceleration level as
\begin{IEEEeqnarray}{LLL}
    \bnudt_e^* = \bm{J}_{e} (\q) \bnudt + \dot{\bm{J}}_e (\q) \bnu,
\end{IEEEeqnarray}
where $\bm{J}_e (\q) \in \mathbb{R}^{6 \times (6+n_j)}$ is the EE task Jacobian. We impose that the EE pose task does not affect the reduced attitude task, by projecting it in its null-space as follows:
\begin{IEEEeqnarray}{LLL}
    \bnudt^\ast_e &= \bm{J}_e (\q) \bm{N}_{\text{tilt}} \bnudt + \bm{J}_e (\q) \bm{J}^{\dagger}_{\text{tilt}} \dot{\bm{\omega}}_{xy} + \dot{\bm{J}}_e (\q) \bnu,
    \\
    \bm{N}_{\text{tilt}} &\triangleq (\bm{I} - \bm{J}^{\dagger}_{\text{tilt}} \bm{J}_{\text{tilt}}),
\end{IEEEeqnarray}
where $\bm{N}_{\text{tilt}}$ is the null-space projection matrix in the reduced attitude task~\cite{khatib1987}, $\bm{I}$ is the identity matrix with the respected size. The superscript $^\dagger$ denotes the Moore-Penrose inverse.
\subsubsection{Posture Task}
\label{subsection:priority 3}
The third priority task comprises a desired posture of the AM. This task is incorporated as the least square error in the cost objective of the OCP. Notice that this cost is equivalent to projecting the task in the null space of the higher-priority task. 

Finally, we can write down the nominal TSID  QP that solves for the above stack of tasks.      

\begin{Problem}Nominal TSID for Aerial Manipulation
\label{problem:nominal TSID}
\begin{subequations}    
\begin{IEEEeqnarray}{LLL}
\underset{\bm{u},\bm{\dot{\nu}}}{\text{minimize}} \quad ||\bnudt- \bnudt^\ast_{\text{Nom-posture}}||^2_{\bm{W} }
\\
\text{subject to} \nonumber 
\\
\label{eq:dynamics nom}
\quad \bm{M}(\q) \bnudt + \bm{h}(\q,\bnu) = \bm{G}(\q)\ip + \jacobc^{T}\hat{\bm{F}}_c ,
\\
\label{eq:tilt task nom}
\quad \dot{\bm{\omega}}_{xy}^* = \bm{J}_{\text{tilt}} \dot{\bm{\nu}},
\\
\quad\bnudt_e^* = \bm{J}_e (\q)\bm{N}_{\text{tilt}} \bnudt + \bm{J}_e (\q) \bm{J}^{\dagger}_{\text{tilt}} \dot{\bm{\omega}}_{xy}^* +\dot{\bm{J}}_e (\q) \bnu , \label{eq:posture nom}
\\
\label{eq:control limits nom}
\quad\underline{\bm{{u}}} \leq \bm{u} \leq \overline{\bm{{u}}},   
\quad\underline{\bm{{\dot{\nu}}}} \leq \bm{\dot{\nu}} \leq \overline{\bm{{\dot{\nu}}}},
\end{IEEEeqnarray}
\end{subequations}
\end{Problem}
where the desired nominal posture controller is 
\begin{IEEEeqnarray}{LLL}    
\bnudt^\ast_{\text{Nom-posture}} = \bm{K}_p (\q_\text{des} - \q) -\,\bm{K}_d\,\bm{\nu}.
\end{IEEEeqnarray}
$\bm{W} \succeq 0$ denotes the posture task weight, and $\hat{\bm{F}}_c$ represents the measured contact forces. 
\begin{Remark}
Problems~\eqref{eq:vanilla TSID problem} and~\eqref{problem:nominal TSID} do not account for post-impact velocities, relying instead on fast feedback loops to address impact disturbances during contact establishment, which leads to control spikes as presented in Sec.~\ref{sec:control challenges}. This remains a challenging research question for robots performing contact-rich tasks, such as loco-manipulation~\cite{Kheddar2023, 9636094}.   
\end{Remark}

To address this drawback in nominal TSID feedback controllers, we leverage the developments in Sec.~\ref{sec:method} to include the impact-robustness residual in the cost function as a remedy to account for post-impact impulses that can be difficult for only feedback to handle. We formulate the following OCP:

\begin{Problem}Impact-Robust TSID for Aerial Manipulation
\label{problem:impact-robust TSID}
\begin{subequations}    
\begin{IEEEeqnarray}{LLL}
\underset{\bm{u},\bm{\dot{\nu}}}{\text{minimize}} \quad ||\bnudt - \bnudt^\ast_{\text{IM-posture}}||^2_{\bm{W}}
\\
\text{subject to} \nonumber 
\\
\eqref{eq:dynamics nom}, \eqref{eq:tilt task nom}, \eqref{eq:posture nom}, \eqref{eq:control limits nom}. \nonumber
\end{IEEEeqnarray}
\end{subequations}
\end{Problem}
The desired impact posture acceleration command becomes:
\begin{IEEEeqnarray}{LLL}    
\label{eq:impact-robust residual}
    \bm{\dot{\nu}}^\ast_{\text{IM-posture}} = \bnudt^\ast_{\text{Nom-posture}} - \bm{K} \nabla_{\bm{q}} H(\q).
    \label{eq:postural_task}
\end{IEEEeqnarray}
Notice that the only difference between the impact-robust TSID problem above and the~\eqref{problem:nominal TSID} is the gradient of the impact-robustness term in the desired posture acceleration. Thanks to this term, the impact-robust TSID controller utilizes the robot's kinematic redundancy to iteratively optimize postures that are robust to anticipated impacts. 
\begin{Remark}
\label{remark:computational overhead}
 Since the impact direction is known a priori (Assumption~\eqref{assumption:impacts}), $\nabla_{\bm{q}}H(\q)$ can be computed offline using an off-the-shelf automatic differentiation tools, which results in a negligible computational overhead online w.r.t. the nominal TSID in problem~\eqref{problem:nominal TSID} (an extra matrix-vector multiplication).   
\end{Remark}

%% file: simulations.tex
\section{Results}\label{sec:sim_sens}
We evaluate our method in the Gazebo physics-based simulator \footnote{https://gazebosim.org/home} using the Open Dynamics Engine (ODE) \footnote{https://www.ode.org/} on a fully actuated hexarotor equipped with a 3-DoF robotic arm. The open-source framework telekyb3 \footnote{https://git.openrobots.org/projects/telekyb3} interfaces the robot software stack with the simulator, incorporating Gazebo plugins to model rotor dynamics via a first-order motor model. To enhance simulation fidelity, we model propeller-induced vibrations on the hexarotor body, which affect the simulated IMU measurements. For controller development, we use the Whole-Body toolbox~\cite{RomanoWBI17Journal} to compute dynamic quantities (e.g., inertial, Coriolis, and gravitational matrices) of the aerial manipulator model. The derivative of the impact-robustness metric with respect to the robot configuration, $\nabla_{\bm{q}} H(\mathbf{q})$, is computed using ADAM \footnote{https://github.com/ami-iit/adam} and CasADi auto-differentiation~\cite{Andersson2019}, which is essential for the impact-robustness residual task in Problem~\ref{problem:impact-robust TSID}. Both nominal and impact-robust TSID whole-body control problems are solved using the open-source QP solver qpOASES~\cite{Ferreau2014}.

\begin{figure}[!t]
        \begin{subfigure}{\columnwidth}
        \begin{minipage}[b]{.32\columnwidth}
        \centering
        \includegraphics[scale=0.043]{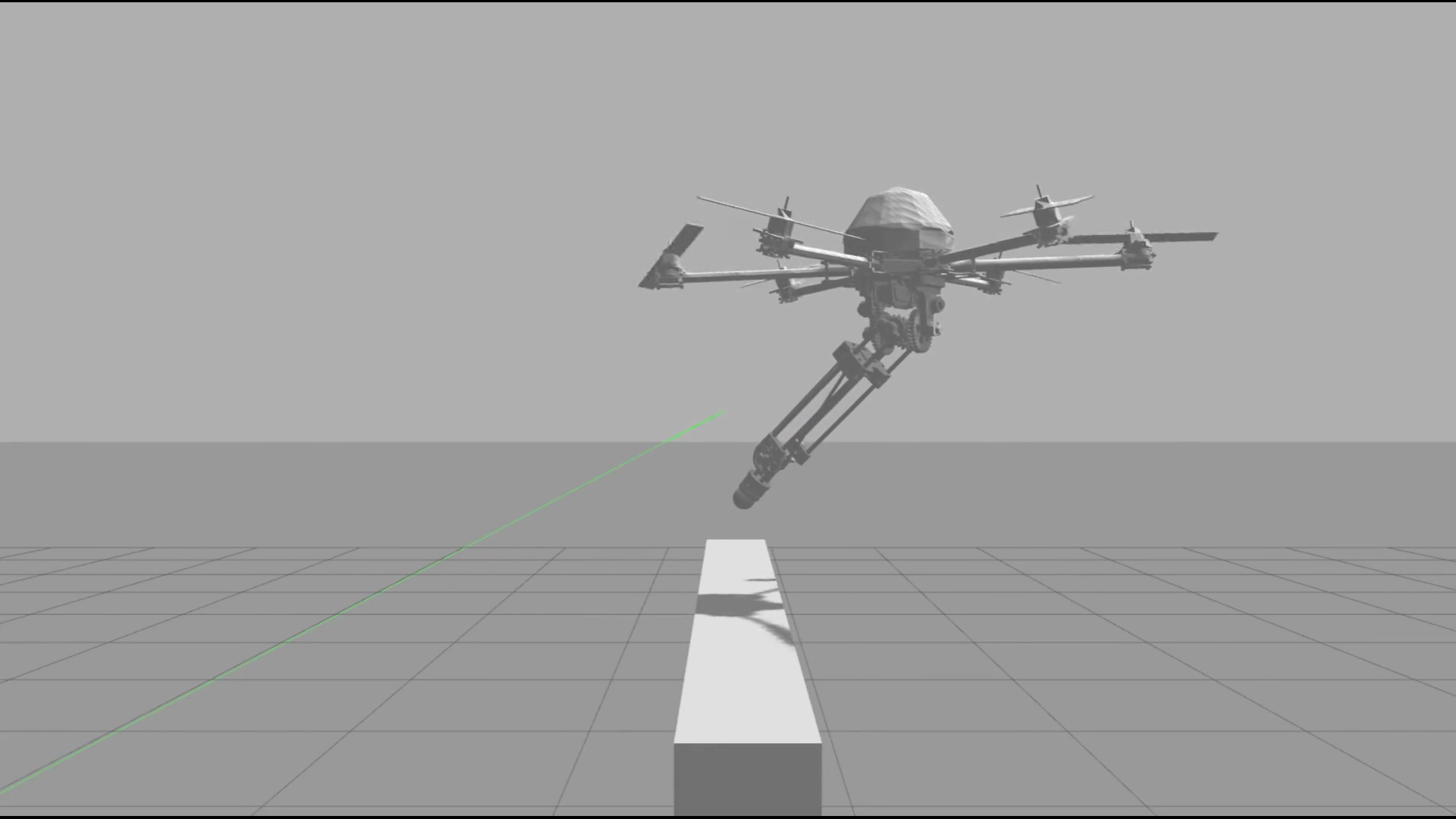}
        \end{minipage}
        \begin{minipage}[b]{.32\columnwidth}
        \centering
        \includegraphics[scale=0.043]{figures/nominal_tsid_vertical_pushing_snaphots/vertical_pushing_nom_1.png}
        \end{minipage}
        \begin{minipage}[b]{.32\columnwidth}
        \includegraphics[scale=0.043]{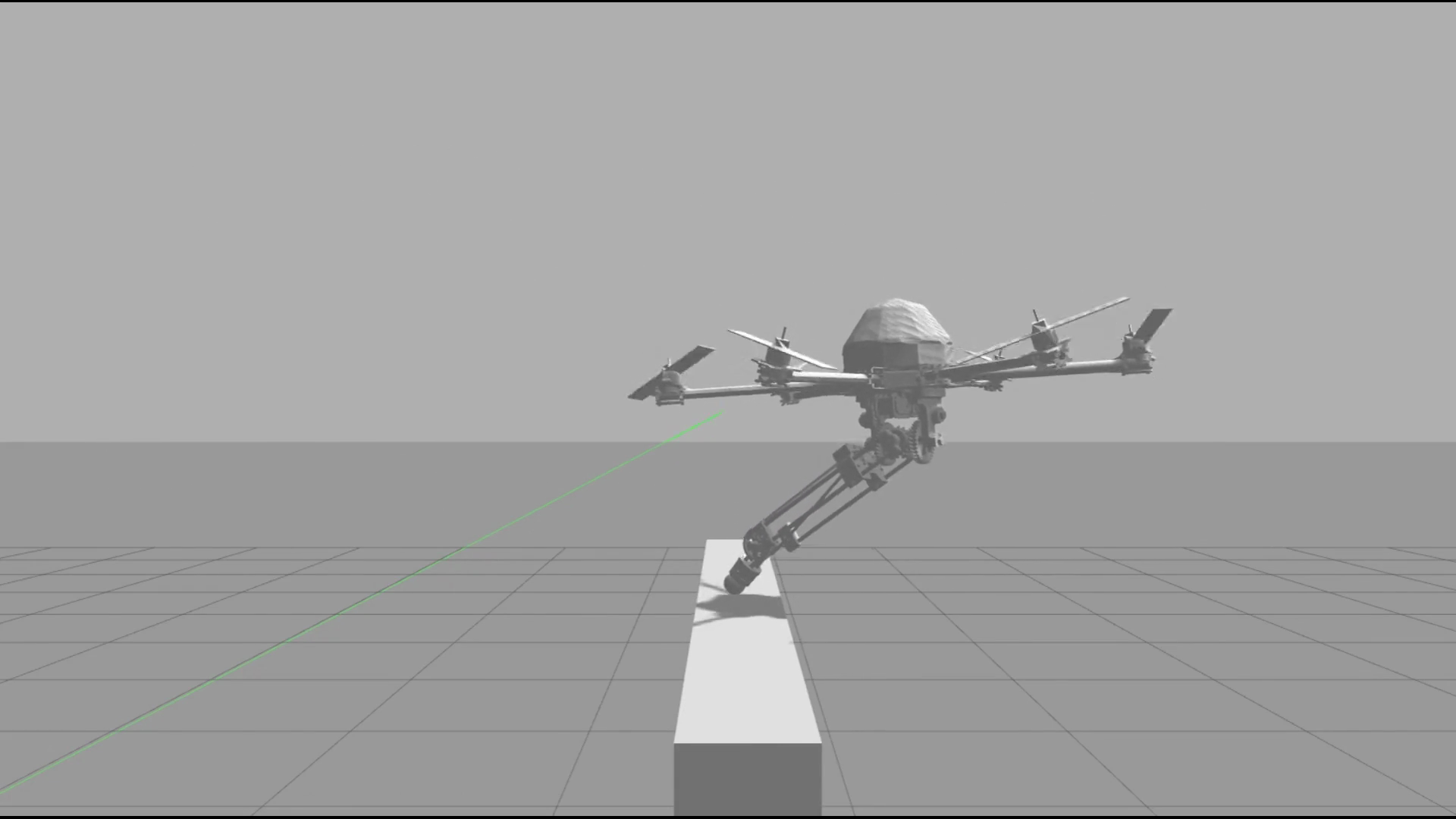}
        \end{minipage}
        \caption{Nominal TSID.}
        \end{subfigure}
        \begin{subfigure}{\columnwidth}
        \begin{minipage}[b]{.32\columnwidth}
        \centering
        \includegraphics[scale=0.043]{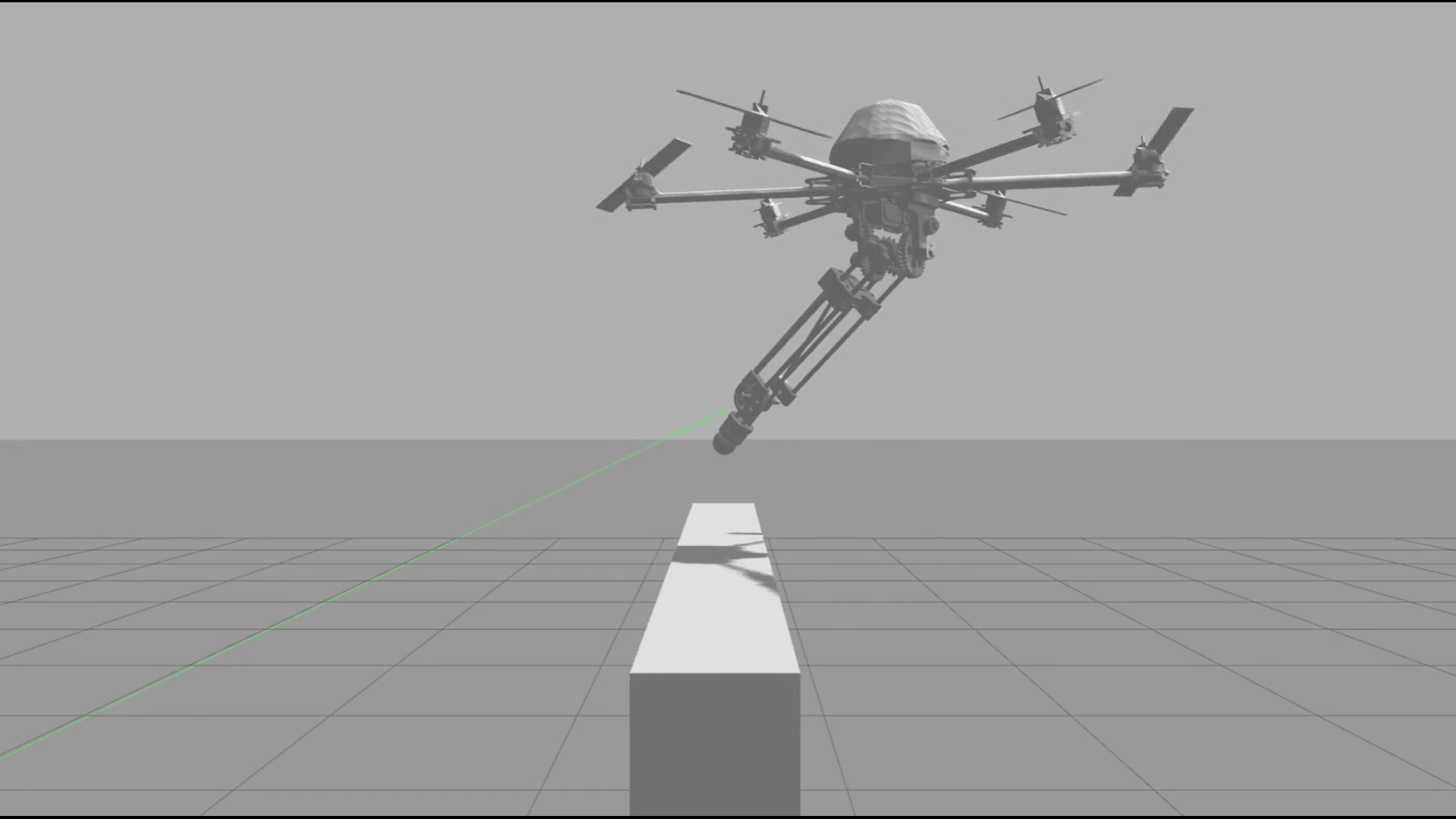}
        \end{minipage}
        \begin{minipage}[b]{.32\columnwidth}
        \centering
        \includegraphics[scale=0.043]{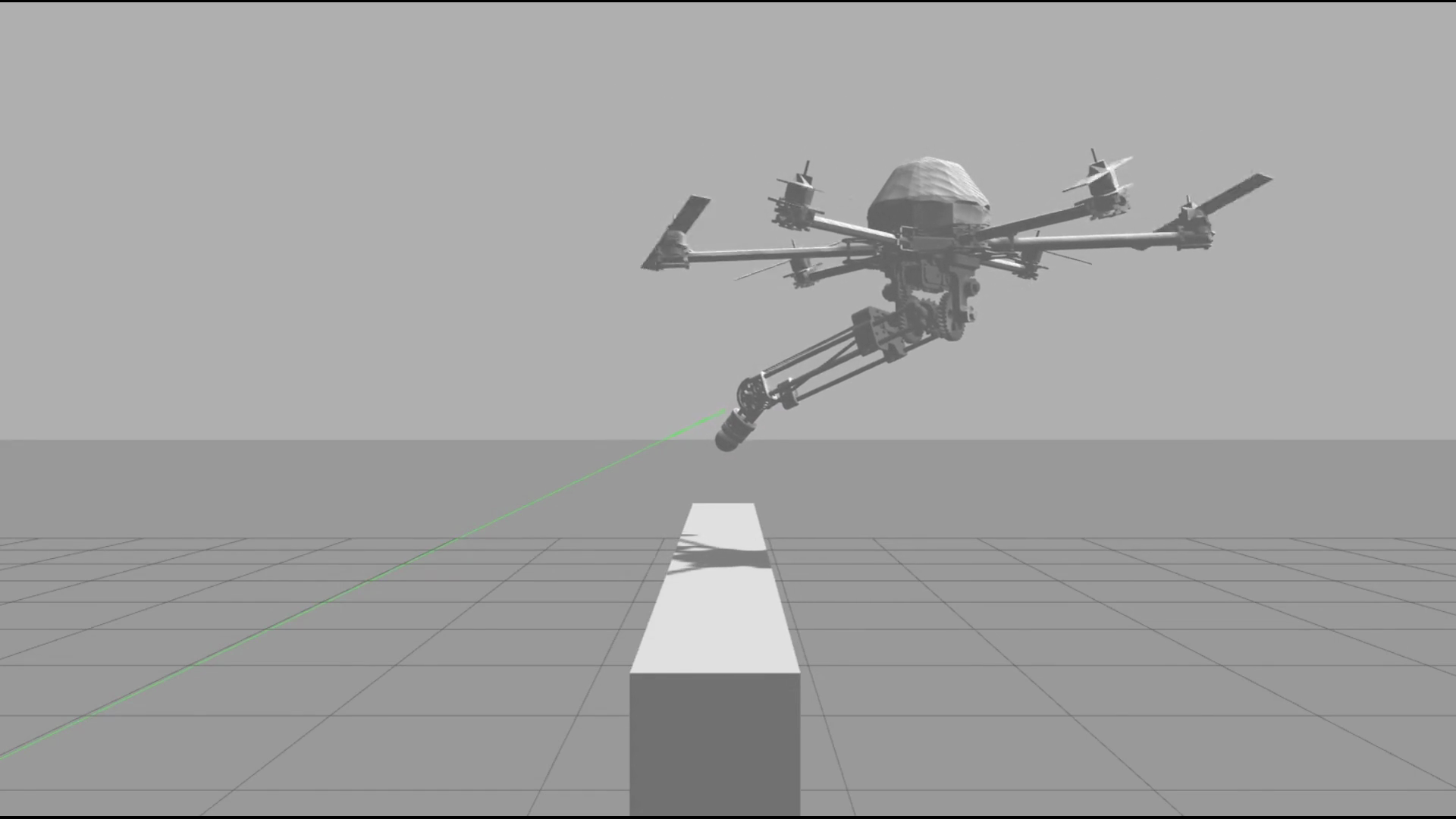}
        \end{minipage}
        \begin{minipage}[b]{.32\columnwidth}
        \includegraphics[scale=0.043]{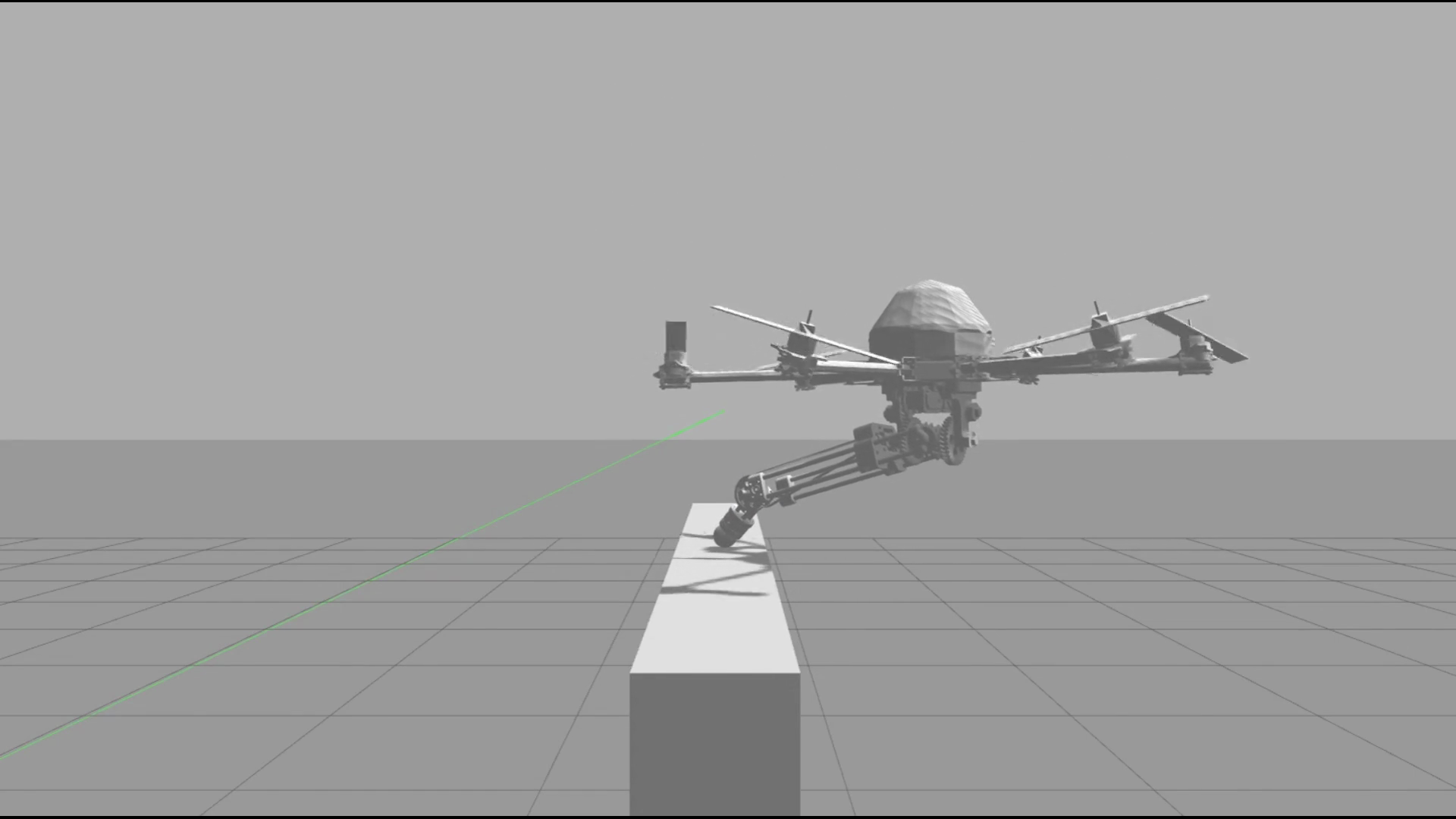}
        \end{minipage}
        \caption{Impact-robust TSID.}
        \end{subfigure}
        \caption{Left to right: snapshots of robot configurations in the vertical contact scenario.}
        \label{fig:vertical_pushing_scenario}
\end{figure}
%
\begin{figure}[!t]
        \begin{subfigure}{\columnwidth}
        \begin{minipage}[b]{.32\columnwidth}
        \centering
        \includegraphics[scale=0.043]{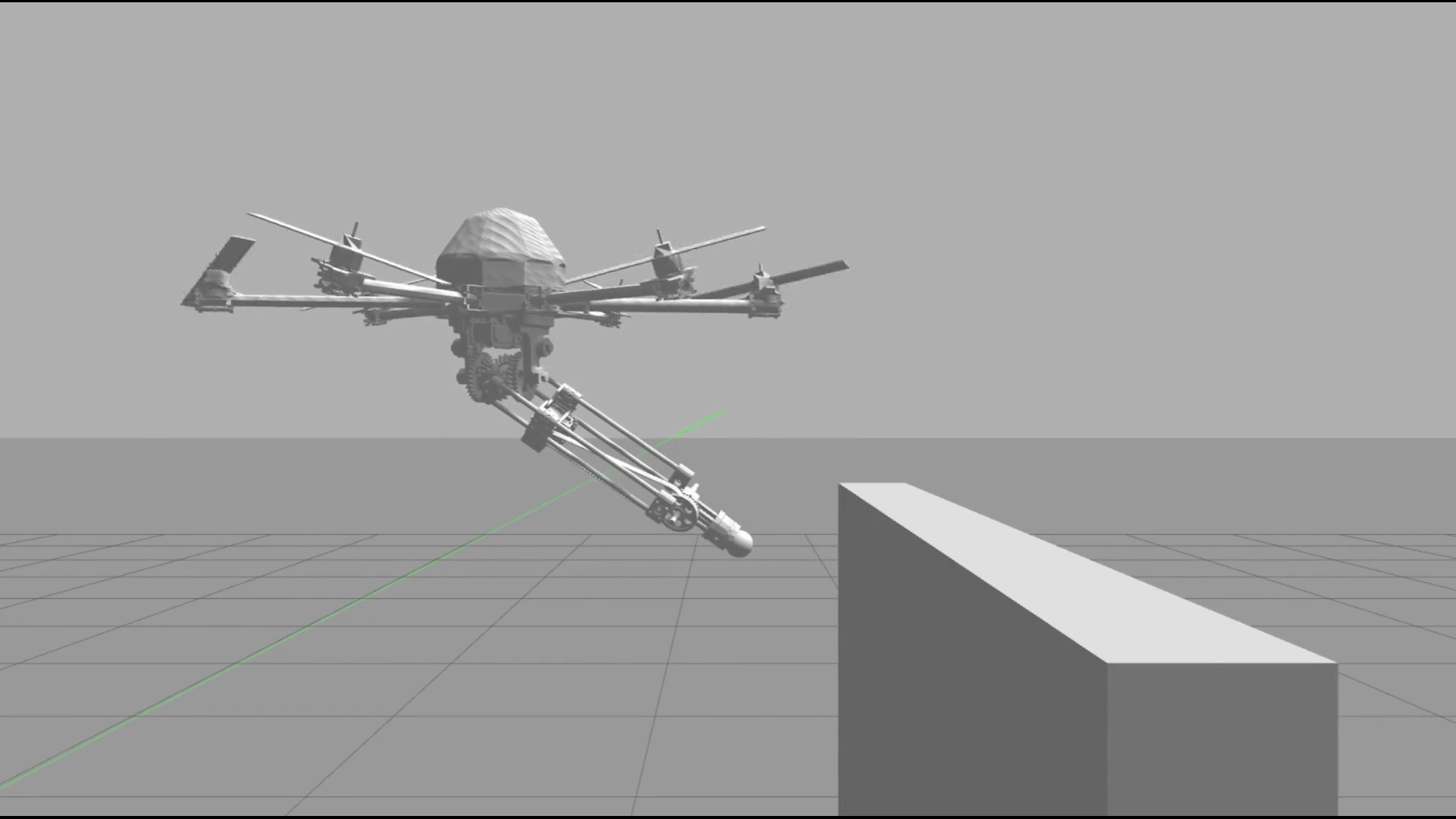}
        \end{minipage}
        \begin{minipage}[b]{.32\columnwidth}
        \centering
        \includegraphics[scale=0.043]{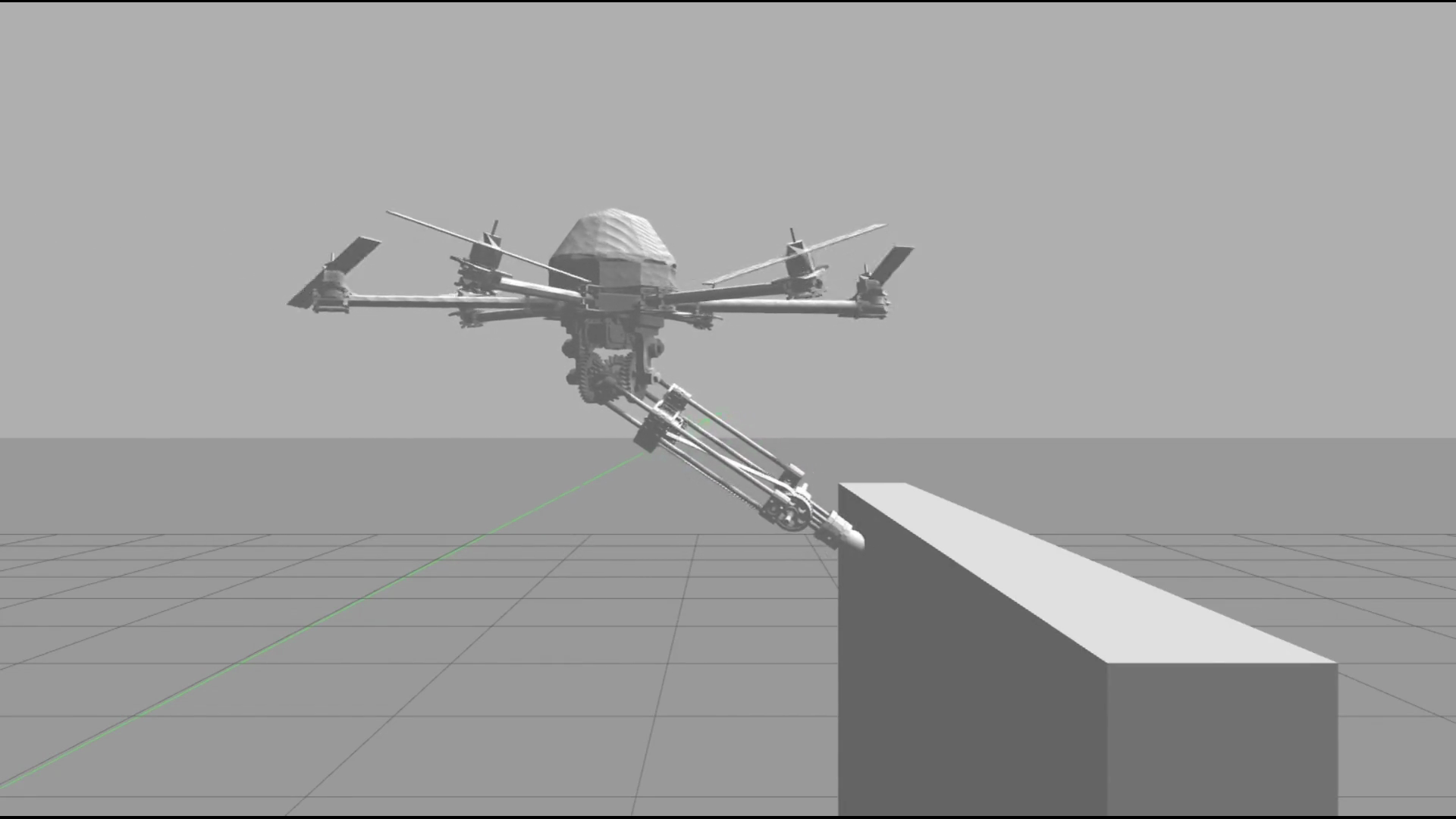}
        \end{minipage}
        \begin{minipage}[b]{.32\columnwidth}
        \includegraphics[scale=0.043]{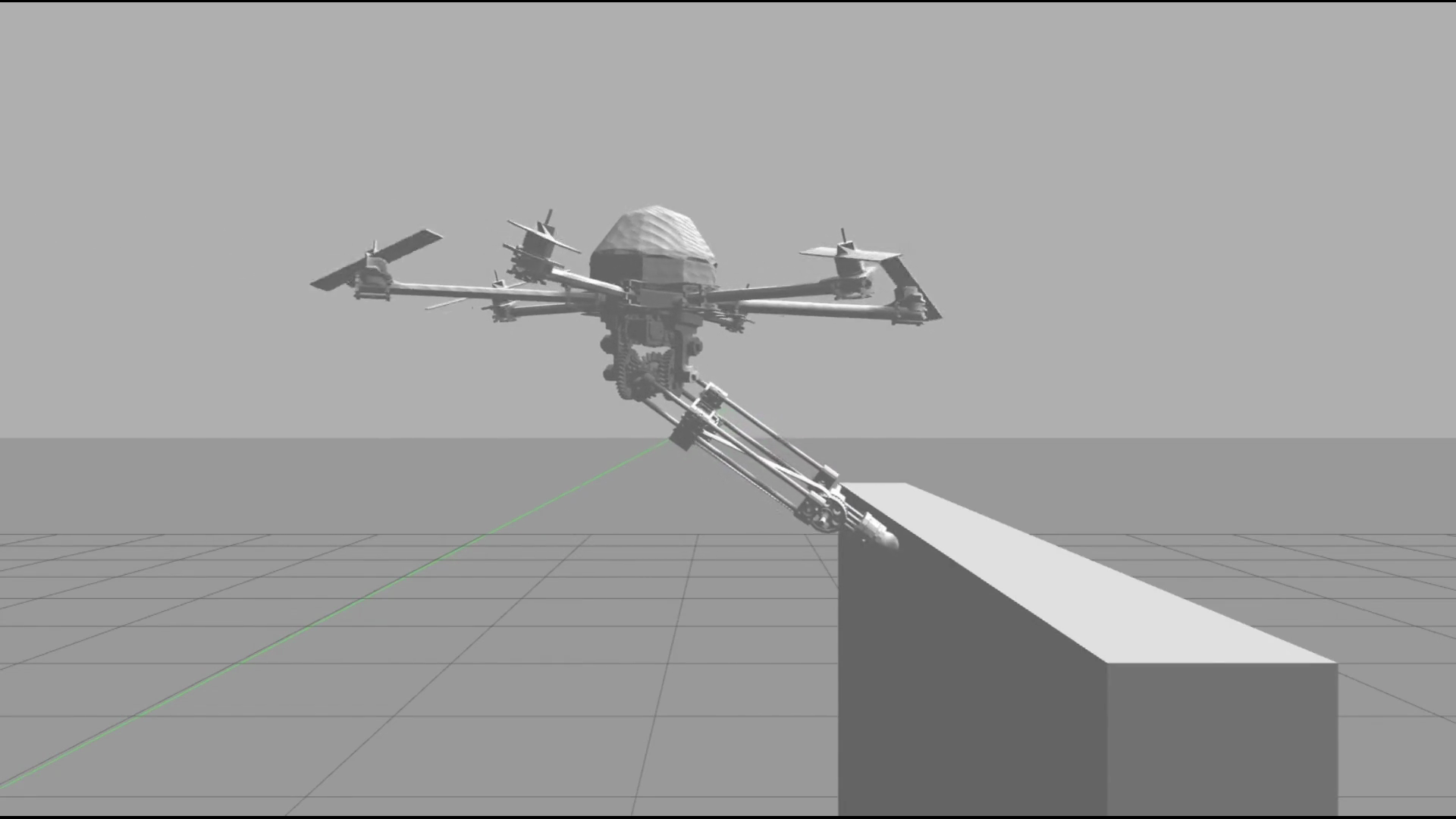}
        \end{minipage}
        \caption{Nominal TSID.}
        \end{subfigure}
        \begin{subfigure}{\columnwidth}
        \begin{minipage}[b]{.32\columnwidth}
        \centering
        \includegraphics[scale=0.043]{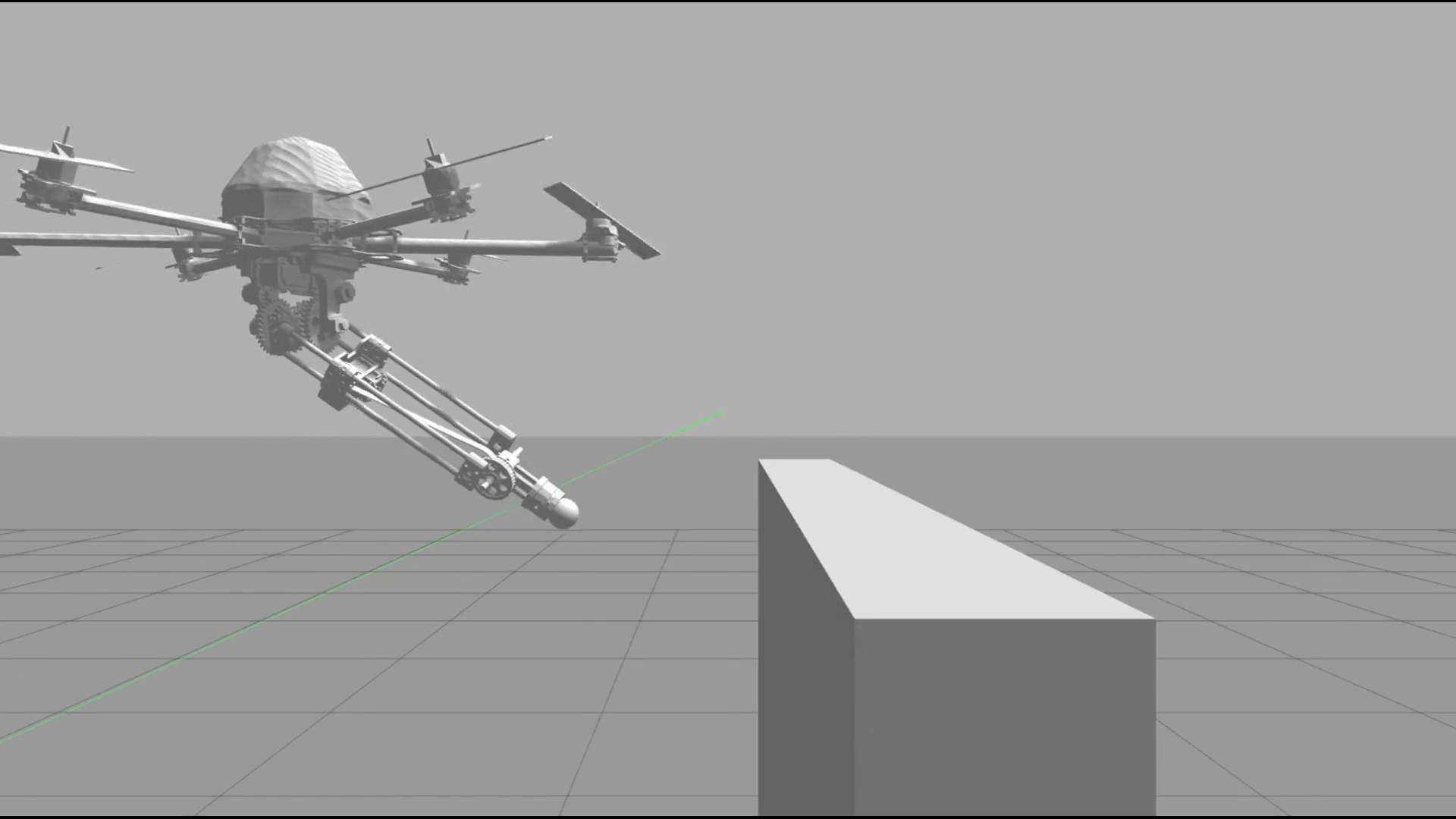}
        \end{minipage}
        \begin{minipage}[b]{.32\columnwidth}
        \centering
        \includegraphics[scale=0.043]{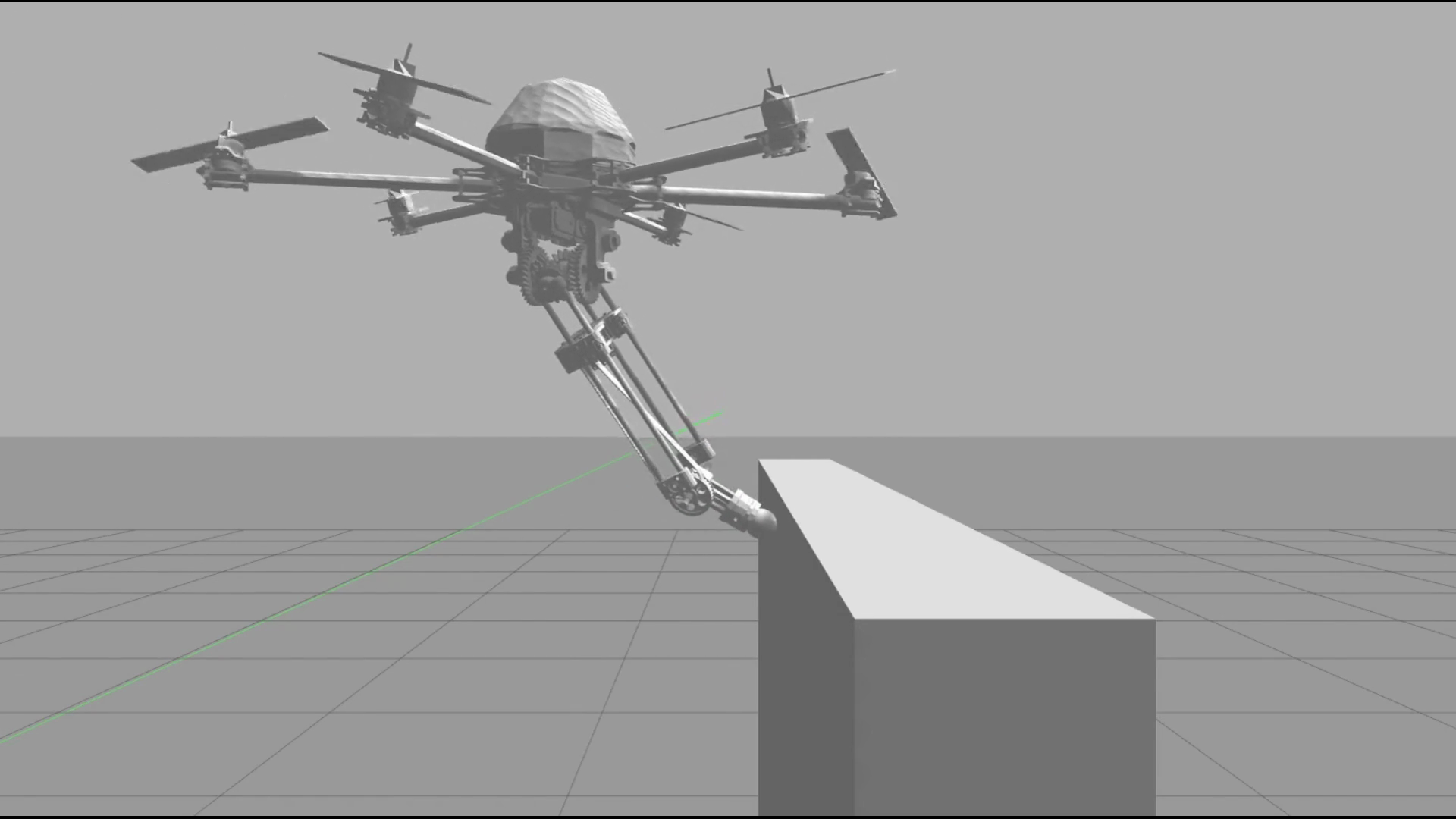}
        \end{minipage}
        \begin{minipage}[b]{.32\columnwidth}
        \includegraphics[scale=0.043]{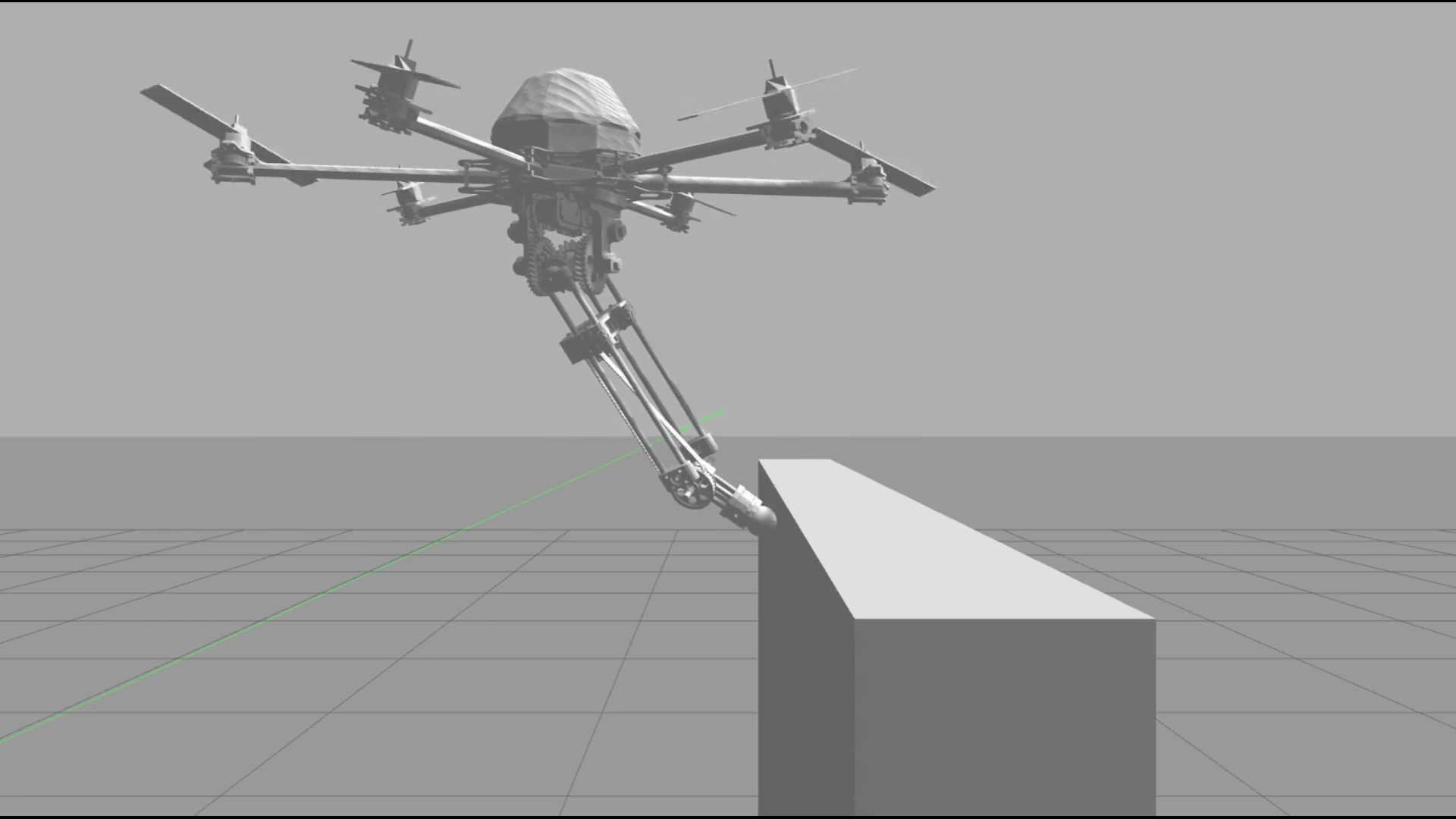}
        \end{minipage}
        \caption{Impact-robust TSID.}
        \end{subfigure}
        \caption{Left to right: snapshots of robot configurations in the lateral contact scenario.}
        \label{fig:lateral_pushing_scenario}
\end{figure}      

\subsection{Aerial Manipulation Scenarios and Robustness metrics}
The AM is tasked with making and breaking contacts repeatedly with a known contact surface normal using its EE, similar to a contact inspection task. In particular, we devise the following contact scenarios:
\begin{itemize}
    \item \textbf{Vertical pushing}: five repeated vertical contacts (Fig.~\ref{fig:vertical_pushing_scenario}).
    \item \textbf{Lateral pushing}: five repeated lateral contacts (Fig.~\ref{fig:lateral_pushing_scenario}). 
\end{itemize}
The robot's whole-body controller is tasked with maintaining zero tilt of the floating base of the hexarotor as its highest priority task (Sec.~\ref{subsection:priority 1}). Moreover, the robot controls a desired EE pose as the second priority (Sec.~\ref{subsection:priority 2}). Finally, the robot is assigned a desired posture as the third priority task (Sec.~\ref{subsection:priority 3}). We compare the robustness of our Impact-robust TSID controller in problem~\eqref{problem:impact-robust TSID} that optimizes for the impact-robustness residual~\eqref{eq:impact-robust residual} against the nominal TSID controller problem~\eqref{problem:nominal TSID} that is impact agnostic. We use the following metrics to quantify the robustness of both controllers during the impact events:
\begin{itemize}
    \item \textbf{Constraint saturation}: We consider the saturation of the control commands during impact events as a critical robustness metric, since constraint saturation leads to unsafe motions. 
     \item \textbf{Motion during impacts}: this metric quantifies the change in robot configuration during impact events, computed as the integral of the $l_2$ norm of the actual robot velocity $\bm{\nu}_a$
     \begin{IEEEeqnarray}{LLL}
        \label{eq:q_total_impact}
         \bm{q}^{\text{total}}_{\text{impact}} = \int_{t=0}^{t=T} ||\bm{\nu}_a(t)| dt,
     \end{IEEEeqnarray}
      where $T$ is the sum of impact durations.  
\end{itemize}
\subsection{Aerial Manipulation Simulations}

\begin{figure}[!t]
\centering
\includegraphics[width=1\columnwidth]{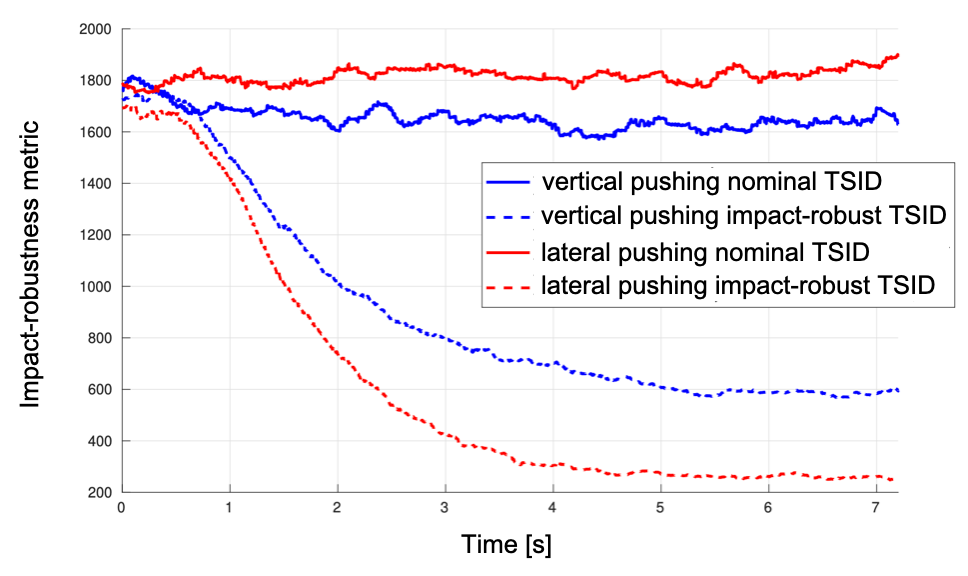}
  \caption{The evolution of the impact-robustness metric $H(\q)$ for the vertical and lateral contact scenarios.}
  \label{fig:reconfig_metric}
\end{figure}

We plot the impact-robustness metric $H(\mathbf{q})$ in Fig.~\ref{fig:reconfig_metric}, which serves as a feedforward term in the desired impact-robustness residual (Eq.~\eqref{eq:impact-robust residual}). Recall that this term drives the robot's whole-body reconfiguration to achieve impact-robust postures. Starting from the same initial configuration for both controllers, the impact-robust TSID controller adjusts the robot's posture to minimize $H(\mathbf{q})$ while maintaining the desired EE pose in both test scenarios. As shown in Fig.~\ref{fig:reconfig_metric}, the impact-robust TSID significantly reduces $H(\mathbf{q})$ for both the vertical and lateral contact scenarios compared to the nominal TSID, proactively accounting for impact disturbances. This reconfiguration can be visualized in Figs.~\ref {fig:vertical_pushing_scenario} and \ref {fig:lateral_pushing_scenario} before the impact event. Thanks to this reconfiguration, it becomes evident that the impact-robust TSID absorbs impacts more gracefully than the nominal TSID (see the multimedia attachment for reference).

\begin{figure}[!t]
\centering
\includegraphics[scale=0.351]{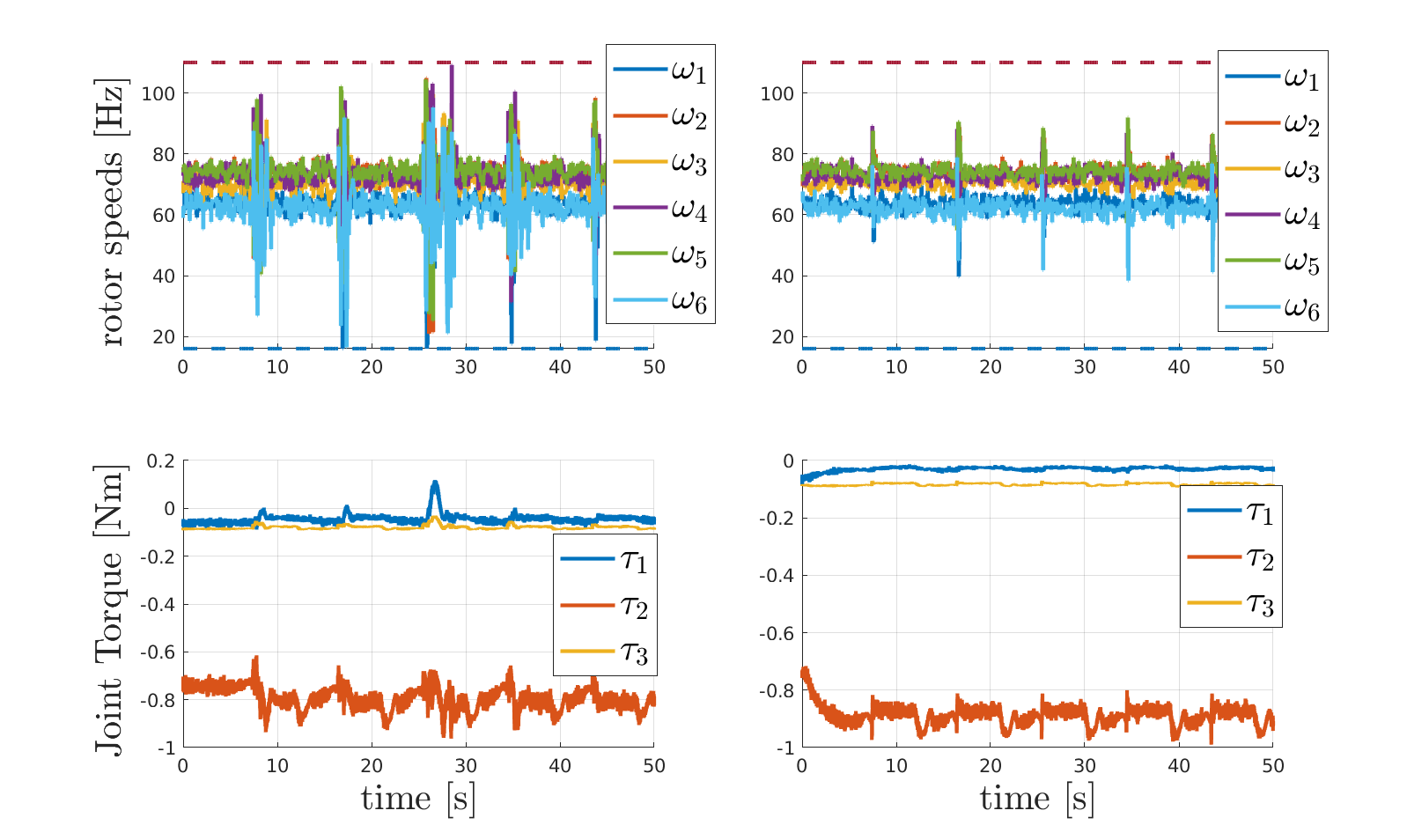}
  \caption{AM's whole-body controls (rotors' angular velocities, and joint torques) for the vertical pushing scenario. Left: nominal TSID controller saturates the angular velocities control limits during impacts. Right: impact-robust TSID controller optimizes smoother and safer control inputs without saturation.} 
  \label{fig:joint_torque_compare}
\end{figure}

\begin{figure}[!t]
\centering
\includegraphics[width=1\columnwidth]{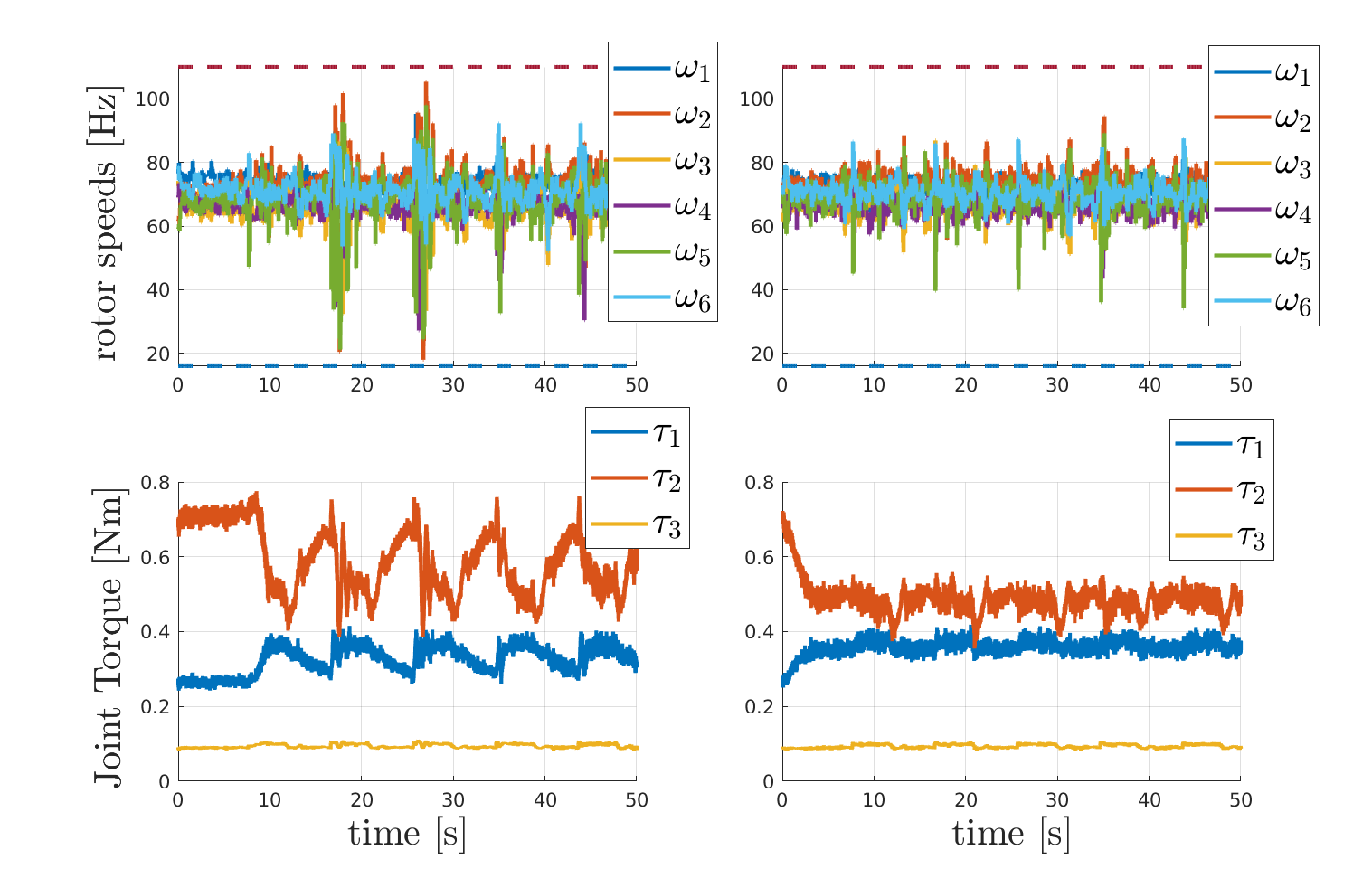}
  \caption{AM's whole-body controls (rotors' angular velocities, and joint torques) for the lateral pushing scenario. Left: nominal TSID controller. Right: impact-robust TSID controller.}
  \label{fig:prop_speeds_compare}
\end{figure}

As a consequence of minimizing the impact-robustness metric $H(\mathbf{q})$ in problem~\eqref{problem:impact-robust TSID}, the impact-robust TSID controller avoids saturating control limits (rotor angular velocities and joint torques) and generates smoother control commands compared to the nominal TSID, as shown in Fig.~\ref{fig:joint_torque_compare} and Fig.~\ref{fig:prop_speeds_compare} for the vertical and lateral contact scenarios, respectively. In contrast, the nominal TSID frequently saturates rotor angular velocity limits and produces aggressive control actions during impacts, since it is agnostic to impacts and relies solely on feedback. Such control saturation compromises robot safety and may cause motion failures in real-world applications.

Finally, we quantified the change in robot configuration during impact events $\bm{q}^{\text{total}}_{\text{impact}}$ in~\eqref{eq:q_total_impact} as our second safety quantification measure. As reported in Table~\ref{table:q_impact}, impact-robust TSID compensates better for impact disturbances, which in turn resulted in less motion during contact events (51\%, and 40.2\%) w.r.t. to nominal TSID for the vertical and lateral pushing scenarios, respectively. The joint configuration profiles are plotted in Fig.~\ref{fig:joint_angle_compare} for the lateral pushing scenario for completeness, which shows the change in joint configurations during the repeated impacts (pink rectangles). As shown in the figure, the robot state varies less during impact for the impact-robust TSID w.r.t. the nominal TSID, thanks to the impact-residual minimization of the posture task~\eqref{eq:impact-robust residual}.

%% file: legged_locomotion.tex
\subsection{Application to contact-rich redundant robots}\label{sec:legged}
\begin{figure}
    \begin{subfigure}[!t]{\columnwidth}
        \begin{minipage}[b]{.23\columnwidth}
        \centering
        \includegraphics[scale=0.112]{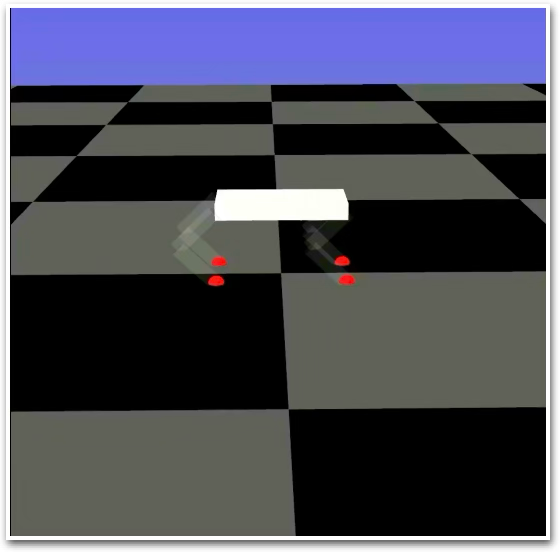}
        \end{minipage}
        \begin{minipage}[b]{.23\columnwidth}
        \includegraphics[scale=0.112]{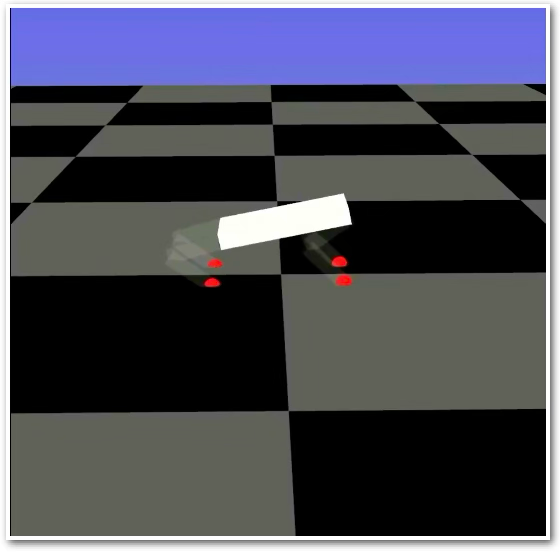}
        \end{minipage}
        \begin{minipage}[b]{.23\columnwidth}
        \centering
        \includegraphics[scale=0.112]{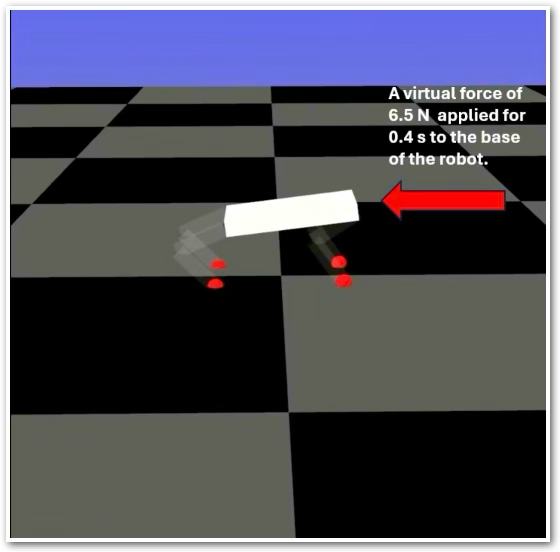}
        \end{minipage}
        \begin{minipage}[b]{.23\columnwidth}
        \centering
        \includegraphics[scale=0.112]{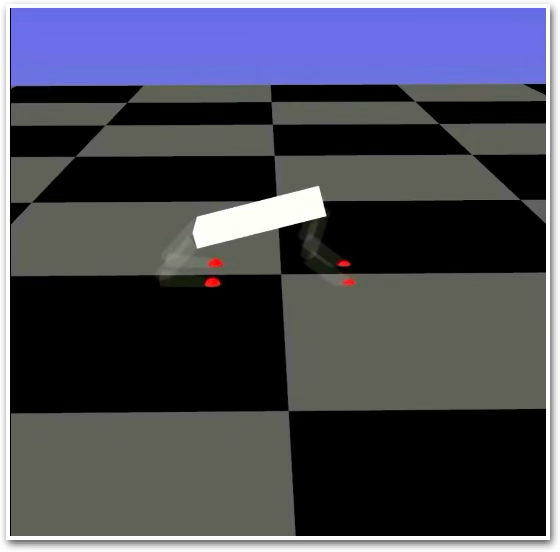}
        \end{minipage}
        \caption{Impact-robust TSID moves less during the impact event.}
        \end{subfigure}
        \begin{subfigure}{\columnwidth}
        \begin{minipage}[b]{.23\columnwidth}
        \centering
        \includegraphics[scale=0.112]{figures/nominal_tsid_solo_snapshots/nominal_tsid_1.png}
        \end{minipage}
        \begin{minipage}[b]{.23\columnwidth}
        \includegraphics[scale=0.112]{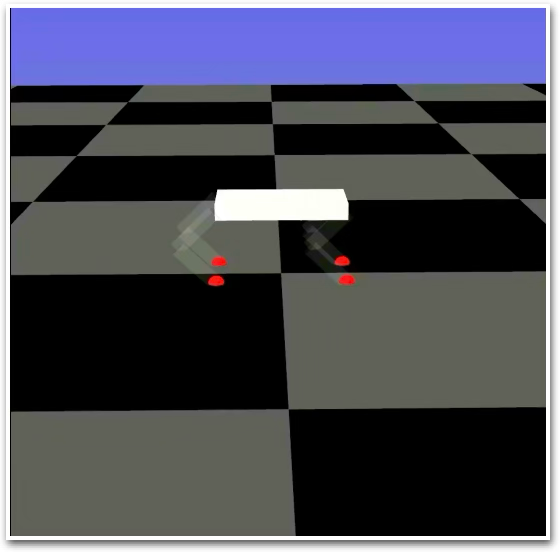}
        \end{minipage}
        \begin{minipage}[b]{.23\columnwidth}
        \centering
        \includegraphics[scale=0.112]{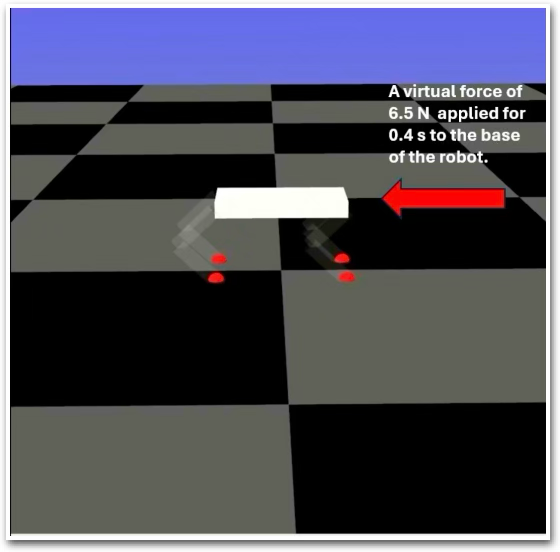}
        \end{minipage}
        \begin{minipage}[b]{.23\columnwidth}
        \centering
        \includegraphics[scale=0.112]{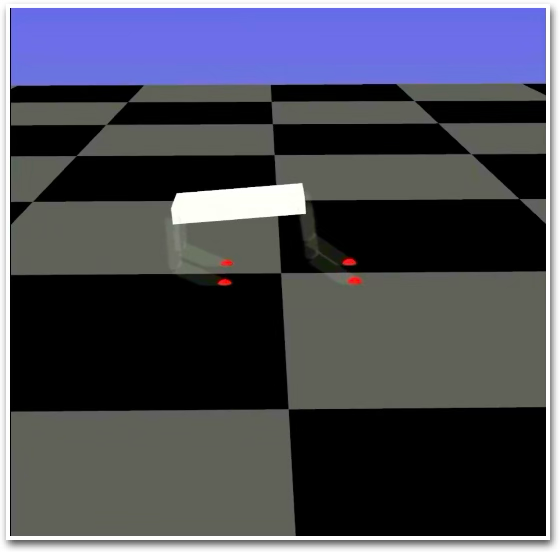}
        \end{minipage}
        \caption{Nominal TSID moves more during the impact event.}
        \end{subfigure}
        \caption{Snapshots of Solo reacting to an impulse force disturbance (red arrow) at the base of $6.5$ N for $400$ ms.}
        \label{fig:solo snapshots}
\end{figure}

 We complement our results with additional numerical simulations performed on highly dimensional multi-contact scenarios for legged robots. We test our approach on the humanoid robot Talos~\cite{stasse2017} and the open-source quadruped robot Solo~\cite{grimminger2020}. Such simulations aim to emphasize the role kinematic redundancy plays in improving impact robustness, which becomes more evident for such systems. Due to the high dimensionality of these systems and the multi-contact nature, we added the following tasks to both the nominal TSID problem~\eqref{problem:nominal TSID} and the impact-robust TSID problem~\eqref{problem:impact-robust TSID}, which are essential for stabilizing the robot.
\begin{itemize}
    \item Center of mass position task: To regulate the configuration of the floating base.  
    \item Friction pyramid constraints: We additionally optimize the contact forces to give more DoFs to the solver, and constrain them to be inside the friction pyramids to ensure dynamic feasibility of the contact forces.
    \item Non-holonomic contact tasks: To constrain the kinematic motion at the robots' end effectors in contact. 
    \item A posture task: to regulate a desired robot posture in the joint space.
\end{itemize}
\begin{table}
\begin{tabular}{ |c|c|c| } 
 \hline
 Test Scenario & $\bm{q}^{\text{total}}_{\text{impact}}$ nominal TSID & $\bm{q}^{\text{total}}_{\text{impact}}$ impact-robust TSID  \\ 
 \hline
 Vertical push  & $10.4$ & $\bm{5.1}$ \\ 
\hline
 Lateral push  & $9.2$ & $\bm{5.5}$ \\ 
\hline
Quadruped  &  $0.411$ & $\bm{0.342}$ \\ 
\hline
Humanoid  &  $0.305$ & $\bm{0.166}$ \\ 
\hline
\end{tabular}
\caption{Comparison of the integral of the $l_2$ norm of the joint velocities over impact duration $\bm{q}^{\text{total}}_{\text{impact}}$.}
\label{table:q_impact}
\end{table}
%
\begin{figure}[!t]
\centering
\includegraphics[width=1\columnwidth]{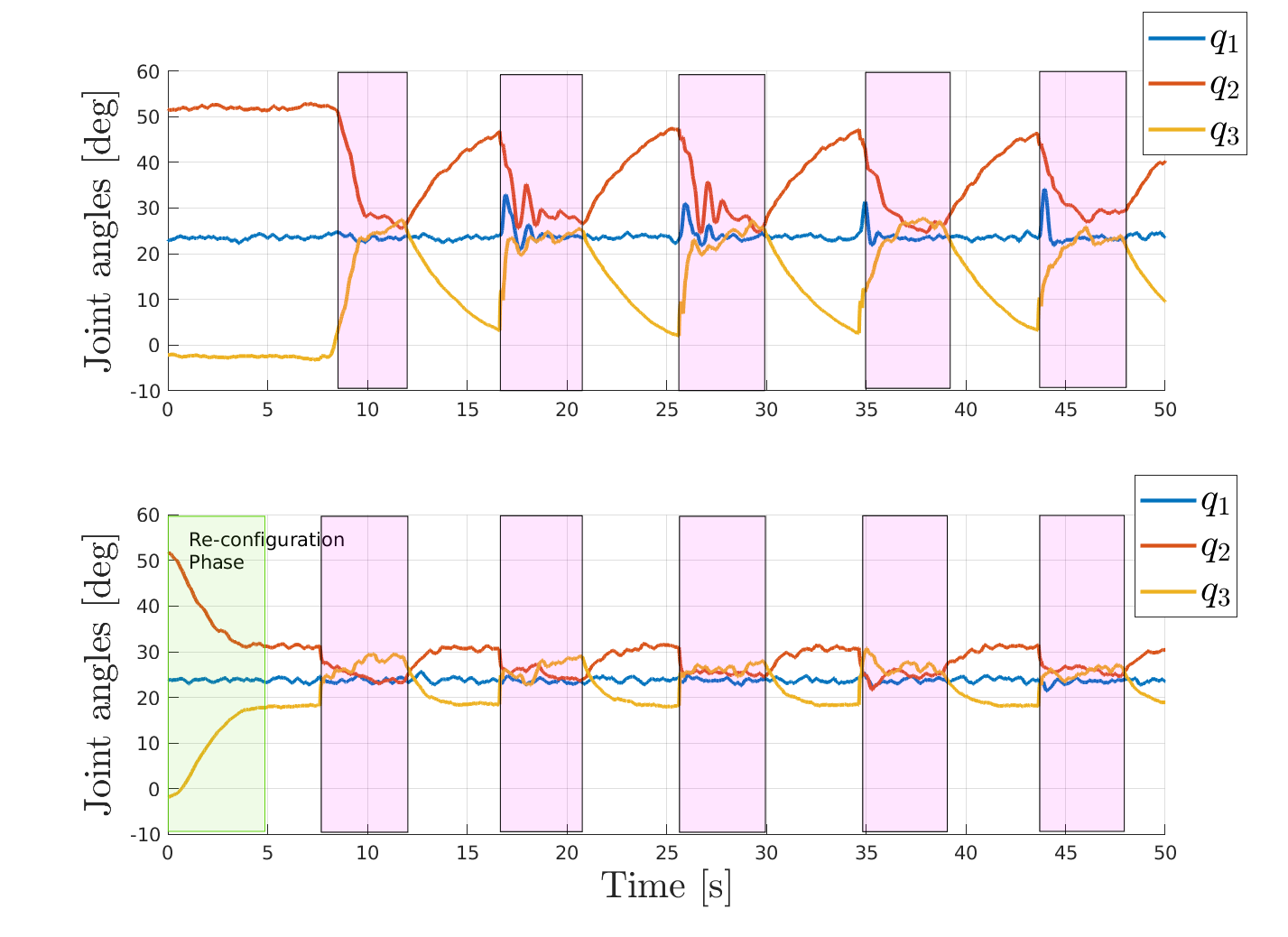}
  \caption{Evolution of the joint angles profiles during impact events (pink rectangles). Top: nominal TSID. Bottom: impact-robust TSID (ours).}
  \label{fig:joint_angle_compare}
\end{figure}
Similarly, we compare our impact-robust TSID approach to the nominal TSID for disturbance compensation for stabilizing an initial configuration of the robot. First, we simulate an impulse force of $6.5$ N for $400$ ms in the sagittal direction of the base of the quadrupedal robot Solo as shown in Fig.~\ref{fig:solo snapshots}. This is an interesting case since this version of Solo legs has only $8$ DoF, which limits the base rotation around the sagittal axis. Despite such a kinematic disadvantage, the impact-robust TSID controller was able to reconfigure the base upwards along the tangential direction to compensate for the force in the sagittal direction (check top row Fig.~\ref{fig:solo snapshots}). Similar to the AM, we report how much the norm of the robot configuration changed during impact events $\q^{\text{total}}_{\text{impact}}$ for both controllers, as shown in Table~\ref{table:q_impact}. Thanks to the impact-robust residual that drives such reconfiguration before impact~\eqref{eq:impact-robust residual}, the impact-robust TSID controller allowed the robot to move less ($16.8\%$ w.r.t. the nominal TSID).

This metric is even more evident in the humanoid robot scenario due to the high kinematic redundancy of Talos. This time we simulate a higher impact of $32$ N for $400$ ms. As shown in Fig.~\ref{fig:talos snapshots}, the impact-robust TSID used the robot's kinematic redundancy as an advantage to reconfigure its upper-body to better compensate for the impact disturbance before the impact. On the other hand, the nominal TSID maintained its initial configuration as it is agnostic to the impact event. $\q^{\text{total}}_{\text{impact}}$ is even more evident here, resulting in $45.6\%$ w.r.t. the nominal TSID controller. This shows that impact-robust TSID can be used as a complementary robustness tool that is easily extendable to highly dimensional multi-contact systems. 
%
%

%% file: discussion.tex
\section{Discussion}
In this section, we outline the strengths and weaknesses of our method and compare it with state-of-the-art approaches. A key strength of our method is its implementation simplicity. By leveraging the rigid impact model residual, the impact-robustness metric is incorporated as an additional feedforward term within the desired posture acceleration task. 
Our method addresses impact discontinuities in a closed-loop feedback manner within a hierarchical TSID, uilizing gradient descent on the impact residual. In contrast, reference spreading handles this issue offline by smoothly blending pre- and post-impact reference states, which does not reduce the actual impulse but rather smooths the controller's tracking target. 

Another key advantage of our method is that it's not conservative. Different from the work of~\cite{Kheddar2023} that enforces extra hard constraints to avoid infeasibilities of the post-impact acceleration, impact-robust TSID tackles the problem by finding impact-robust postures softly through a task residual in the cost prior to the impact. At the same time, a limitation of our method arises when the robot's redundancy is exhausted, as the impact-robustness task is mapped to the null space of higher-priority tasks. Moreover, we differentiate our method from the work of~\cite{9636094}, which takes a robust control approach by mapping the robust controls to an impact-invariant space. Despite being agnostic to impact magnitude, direction, and timing, this approach can be quite conservative and becomes hard to scale for highly dimensional systems like full humanoid robots. Our approach, on the other hand, assumes an impact direction as a trade-off for performance. Thanks to this assumption, impact-robust TSID does not add a computational overhead online w.r.t nominal TSID as the gradient of the impact-robustness term is computed offline.

Finally, we view this work as a complementary robustness tool that can be used in combination with the state-of-the-art methods as~\cite{9867812, Kheddar2023, 9636094, wensing2024}.




%% file: conc.tex
\begin{figure}[!t]
    \begin{subfigure}[t]{\columnwidth}
        \begin{minipage}[b]{.23\columnwidth}
        \centering
        \includegraphics[scale=0.112]{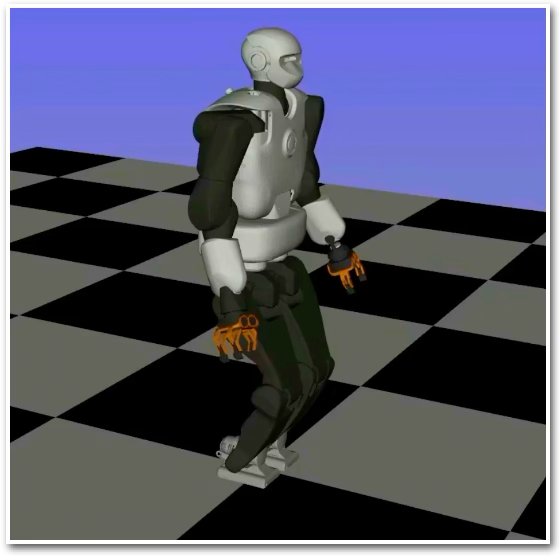}
        \end{minipage}
        \begin{minipage}[b]{.23\columnwidth}
        \includegraphics[scale=0.112]{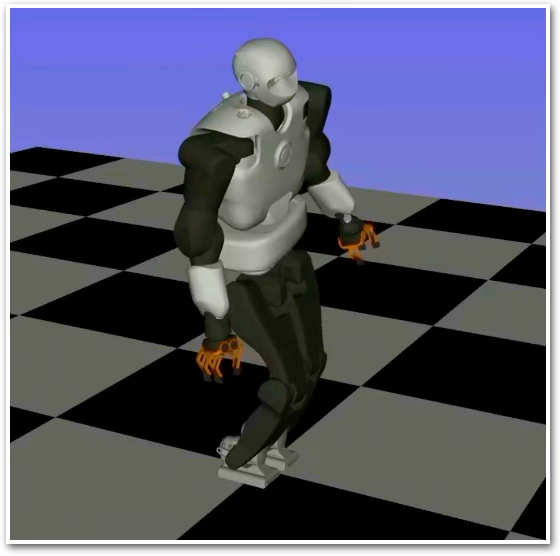}
        \end{minipage}
        \begin{minipage}[b]{.23\columnwidth}
        \centering
        \includegraphics[scale=0.112]{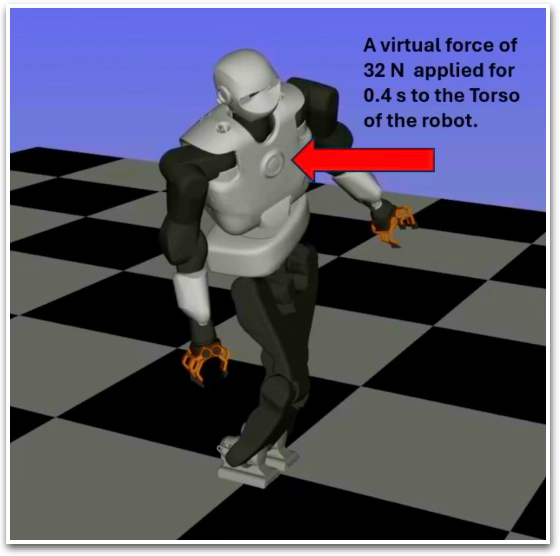}
        \end{minipage}
        \begin{minipage}[b]{.23\columnwidth}
        \centering
        \includegraphics[scale=0.112]{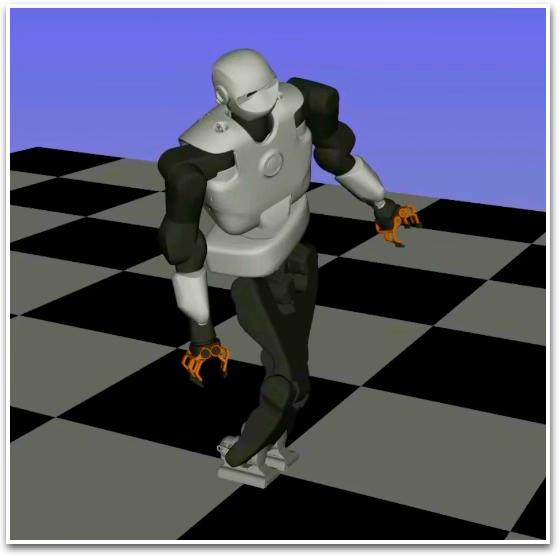}
        \end{minipage}
        \subcaption{Impact-robust TSID optimizes for a posture with lower upper-body momentum to compensate for the impact.}
        \end{subfigure}
        \begin{subfigure}{\columnwidth}
        \begin{minipage}[b]{.23\columnwidth}
        \centering
        \includegraphics[scale=0.112]{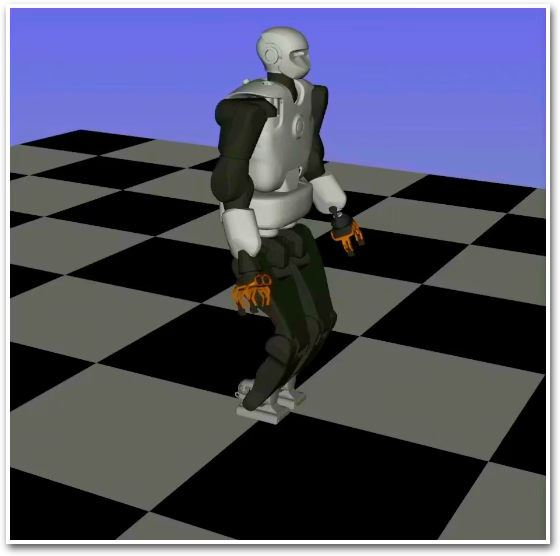}
        \end{minipage}
        \begin{minipage}[b]{.23\columnwidth}
        \includegraphics[scale=0.112]{figures/nominal_tsid_talos_snapshots/nominal_tsid_1.png}
        \end{minipage}
        \begin{minipage}[b]{.23\columnwidth}
        \centering
        \includegraphics[scale=0.112]{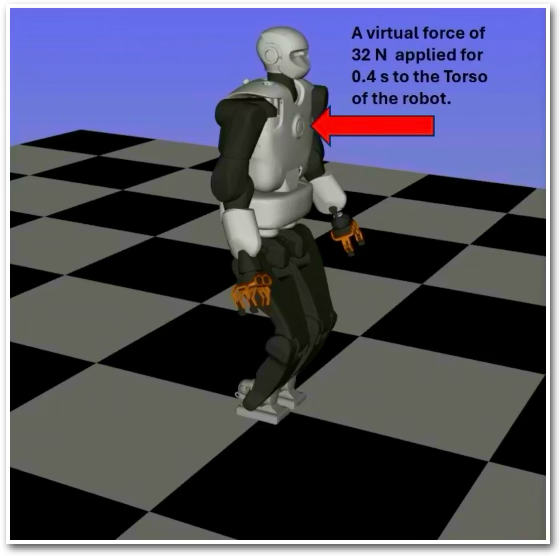}
        \end{minipage}
        \begin{minipage}[b]{.23\columnwidth}
        \centering
        \includegraphics[scale=0.112]{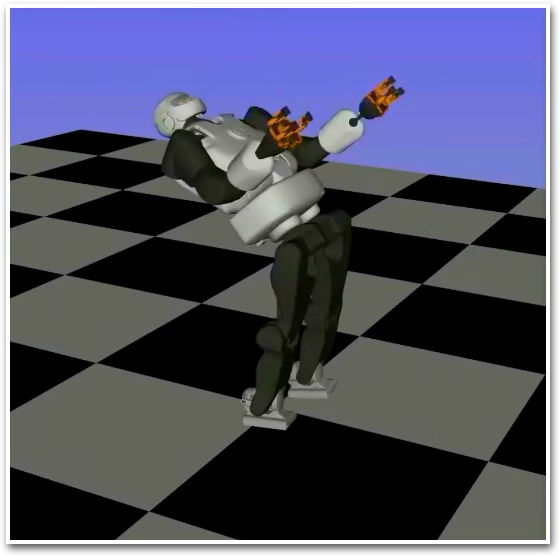}
        \end{minipage}
        \subcaption{Nominal TSID exerts higher upper-body angular momentum to compensate for the impact.}
        \end{subfigure}
        \caption{Snapshots of Talos reacting to an impulse force disturbance (red arrow) at the torso of $32$ N for $400$ ms.}
        \label{fig:talos snapshots}
\end{figure}        
\section{Conclusions}\label{sec:conclusion}
In this work, we presented a novel approach to optimize impact-robust postures as a task residual inside TSID for kinematically redundant robots. Our results show that the optimized posture effectively reduces spikes during impact events, which avoids control saturations and infeasibilities. Consequently, this leads to smoother robot state profiles (joint positions and velocities) that minimize robot motion during impacts, resulting in safer and more stable robot motions. Our method complements state-of-the-art methods in impact handling whenever kinematic redundancy is not exhausted, which is useful for highly dimensional multi-contact robotic systems. Future work includes extending this approach to consider uncertainty in the contact normal and applying our approach to real robot experiments.